\documentclass[lettersize,journal]{IEEEtran}
\usepackage{amsmath,amsfonts}
\usepackage{algorithm}
\usepackage{array}
\usepackage[caption=false,font=normalsize,labelfont=sf,textfont=sf]{subfig}
\usepackage{textcomp}
\usepackage{url}
\usepackage{verbatim}
\usepackage{graphicx}
\usepackage{cite}

\usepackage{algpseudocode}
\usepackage{newfloat}
\usepackage{listings}
\usepackage{multirow}
\usepackage{bibentry}
\usepackage{float}
\usepackage{dblfloatfix}
\usepackage{xcolor}
\usepackage{soul}

\usepackage{graphicx}
\usepackage{bbm}
\usepackage{mathtools}
\usepackage{enumitem}
\usepackage{amsmath}

\begin{document}

\title{Multi-Layer Personalized Federated Learning for Mitigating Biases in Student Predictive Analytics} 

\author{Yun-Wei Chu, Seyyedali Hosseinalipour, Elizabeth Tenorio, Laura Cruz, \\ Kerrie Douglas, Andrew S. Lan, Christopher G. Brinton

\thanks{Y. Chu and C. Brinton are with the Elmore Family School of Electrical and Computer Engineering, Purdue University, IN, USA. e-mail: \{chu198, cgb\}@purdue.edu}
\thanks{S. Hosseinalipour is with the Department of Electrical Engineering, University at Buffalo (SUNY), NY, USA. email: alipour@buffalo.edu}
\thanks{L. Cruz is with the Department of Engineering Education, University of Florida, FL, USA. e-mail: cruzcastrol@ufl.edu}
\thanks{K. Douglas is with the School of Engineering Education, Purdue University, IN, USA. e-mail: douglask@purdue.edu}
\thanks{A. Lan is with the Manning College of Information and Computer Sciences, University of Massachusetts Amherst, MA, USA. email: andrewlan@cs.umass.edu}
\thanks{An abridged version \cite{https://doi.org/10.48550/arxiv.2208.01182} of this paper has been published in the 31st ACM International Conference on Information and Knowledge Management.}

}

\markboth{Journal of \LaTeX\ Class Files,~Vol.~14, No.~8, August~2023}%
{Shell \MakeLowercase{\textit{et al.}}: A Sample Article Using IEEEtran.cls for IEEE Journals}


\maketitle

\begin{abstract}
% Traditional learning-based approaches to student modeling (e.g., predicting grades based on measured activities) generalize poorly to underrepresented/minority student groups due to biases in data availability.
Conventional methods for student modeling, which involve predicting grades based on measured activities, struggle to provide accurate results for minority/underrepresented student groups due to data availability biases.
In this paper, we propose a Multi-Layer Personalized Federated Learning (MLPFL) methodology that optimizes inference accuracy over different layers of student grouping criteria, such as by course and by demographic subgroups within each course.
In our approach, personalized models for individual student subgroups are derived from a global model, which is trained in a distributed fashion via meta-gradient updates that account for subgroup heterogeneity while preserving modeling commonalities that exist across the full dataset. 
The evaluation of the proposed methodology considers case studies of two popular downstream student modeling tasks, knowledge tracing and outcome prediction, which leverage multiple modalities of student behavior (e.g., visits to lecture videos and participation on forums) in model training.
% Experiments on three real-world datasets from online courses demonstrate that our approach obtains substantial improvements over existing student modeling baselines in terms of increasing the average and decreasing the variance of prediction quality across different student subgroups.
Experiments on three real-world online course datasets show significant improvements achieved by our approach over existing student modeling benchmarks, as evidenced by an increased average prediction quality and decreased variance across different student subgroups.
Visual analysis of the resulting students' knowledge state embeddings confirm that our personalization methodology extracts activity patterns clustered into different student subgroups, consistent with the performance enhancements we obtain over the baselines.
\end{abstract}

\vspace{-0.15in}

\vspace{-0.1in}
\section{Introduction}\label{sec:introduction}
% \IEEEPARstart{O}{nline} learning~\cite{Zhang2017DynamicKM} has proven to be a critical component of today's educational platforms, highlighted by its significant uptick in usage during the COVID-19 pandemic~\cite{Adedoyin2020Covid19PA}.
% The remote nature of online learning makes it harder for instructors to pay attention to each individual student and provide feedback at a personalized level. 
% This has motivated the investigation of artificial intelligence (AI)-based approaches to delivering personalized instruction and feedback based on measured student progress in online learning activities \cite{reddy2016unbounded,bassen2020reinforcement}.

\IEEEPARstart{O}{nline} learning, underscored by its substantial surge during the COVID-19 pandemic~\cite{Adedoyin2020Covid19PA}, has become a critical component of contemporary educational platforms~\cite{Zhang2017DynamicKM}.
The lack of in-person interaction in online education poses challenges for instructors in giving personalized attention to individual students and delivering tailored feedback.
This need for tailored feedback has motivated the exploration of AI-driven methods for providing personalized guidance and feedback in online learning based on measured student progress in online learning activities~\cite{Chou2017OpenSM}.

\emph{Student modeling} \cite{vanlehn1988student} and its associated research area aims to produce analytics that may inform such personalization efforts. 
% There exist a wide range of student models, from those that analyze (i) student \emph{knowledge}, such as item response theory \cite{van2013handbook} and models for knowledge tracing \cite{corbett1994knowledge}, to those that analyze (ii) student \emph{behavior}, e.g., to detect psychological states \cite{karumbaiah2021using,yang2019active}, discover learning tendencies \cite{chan2021clickstream,yao2021stimuli}, and identify engagement patterns in discussion forums \cite{Vieira2022StudyOT, Sahay2023PredictingLI}. 
% Since these student models are fitted from student data collected from real-world learning platforms, they are inherently prone to any biases that exist in the available data \cite{fan2019gender}. 
% As a result, the research topic of \emph{de-biasing} data-driven student models has gained recent traction; studies have investigated existing algorithmic biases in educational applications \cite{gardner2019evaluating,kizilcec2020algorithmic} and explored how to impose constraints during model training to promote fairness across different student groups \cite{yao2017beyond}.
A broad spectrum of student models have emerged, including those focused on (i) evaluating student \emph{knowledge}, such as item response theory \cite{van2013handbook} and models for knowledge tracing \cite{corbett1994knowledge}, and those addressing (ii) student \emph{behavior}, for instance, to recognize psychological states \cite{yang2019active}, unveil learning tendencies \cite{chan2021clickstream}, and uncover engagement patterns in discussion forums \cite{ Sahay2023PredictingLI}.
Because these student models are constructed using data acquired from actual learning platforms, they are naturally prone to any biases present within the accessible data \cite{fan2019gender}.
As a result, addressing biases in data-driven student models has gained traction, and research has delved into examining inherent algorithmic biases in educational contexts \cite{kizilcec2020algorithmic}. The methodologies explored to mitigate bias inherited from the data include introducing constraints during model training that ensure equitable outcomes among diverse student groups~\cite{yao2017beyond}.

% One common theme of these existing de-biasing works is that they study the bias/fairness aspects of a single \emph{global} student model that is trained on \emph{all} students data~\cite{paquette2020s}.
A shared focus in these prior de-biasing studies is the examination of bias/fairness within a singular \emph{global} student model that encompasses data from \emph{all} students~\cite{paquette2020s}.
This setup is typically effective in AI applications since more data generally leads to improved model fit.
However, this setup ignores the fact that \emph{underrepresented groups} may not be well-captured by a population-level model, resulting in unfair predictions that could severely impact some students~\cite{Buolamwini:Gender:2018,Lahoti:Ifair:2019}.
% On the contrary, training separate \emph{local} models for each student subgroup may not be effective since small subgroups do not have enough data for us to train an accurate model. In this work, we aim to develop a \textit{personalized student modeling methodology} that addresses the data availability challenges across subgroups.
 In contrast, training separate \emph{local} models for each student subgroup might prove ineffective, as smaller subgroups lack sufficient data for accurate model training. This study endeavors to formulate a \textit{personalized student modeling methodology} to tackle the data availability issue challenges across subgroups.

\begin{figure}[t]
    \centering
    \setlength{\abovecaptionskip}{2mm}
    \includegraphics[width=0.9\linewidth]{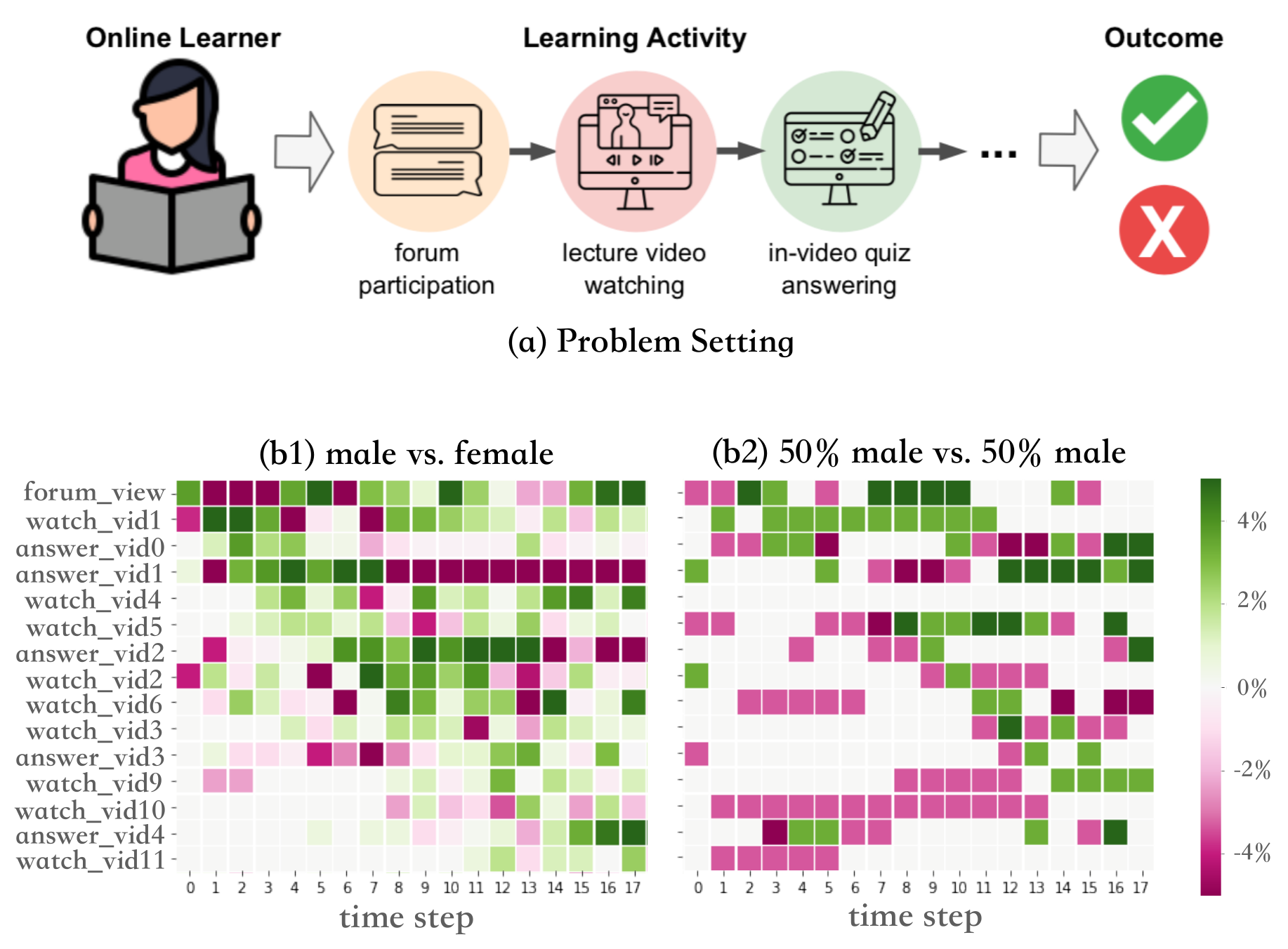}
	 % \caption{(a) Our problem setting of tracing online student learning activity to predict learning outcomes. (b) Heatmaps of differences in learning activities exhibited (b1) between different student subgroups, male vs.\ female, and (b2) within the male subgroup for our QDS dataset.}
  \caption{(a) Our problem setting of tracing online student learning activity to predict learning outcomes. (b) Heatmaps illustrate differences in learning behaviors observed (b1) across distinct student subgroups, such as males versus females, and (b2) within the male subgroup, using our QDS dataset.
}
    \vspace{-0.2in}
	 \label{fig:setting_heat}
\end{figure}

\vspace{-0.15in}
\subsection{Federated Learning and Student Modeling}
% The recently proposed federated deep knowledge tracing (FDKT)~\cite{Wu2021FederatedDK} method is a first step towards coordinating between global and local student models, based on federated learning (FL)~\cite{Konecn2016FederatedLS}, a popular technique for enabling collaboration among models trained on heterogeneous local datasets through periodic global model aggregations.
The recently introduced federated deep knowledge tracing (FDKT) approach \cite{Wu2021FederatedDK} makes an initial attempt to coordinate between global and local student models. It leverages federated learning (FL) \cite{Konecn2016FederatedLS}, a popular technique for enabling collaboration among models trained on heterogeneous local datasets through periodic global model aggregations.

Beyond the conventional FedAvg algorithm~\cite{McMahan2017CommunicationEfficientLO}, research in \textit{Global FL} has introduced diverse aggregation techniques to optimize the obtained global model, in terms of relevant objectives for different learning settings, e.g., ~\cite{Ji2019LearningPN}.
% In particular, FedAtt~\cite{Ji2019LearningPN} adds an attention layer to improve FL in terms of trained model accuracy, while FedAdagrad~\cite{Reddi2021AdaptiveFO} and MOON~\cite{Li2021ModelContrastiveFL} aim to improve the speed of convergence.
However, statistical heterogeneity across local data distributions -- i.e., non-i.i.d. (independent and identically distributed) data -- is an inherent characteristic of most FL applications that presents fundamental challenges to Global FL~\cite{huang2021personalized}. This has motivated \textit{Personalized FL} which aims to redefine the relationship between the local and global models for a better match to individual local datasets~\cite{Kairouz2021AdvancesAO}.
In particular, personalized FL shifts the optimization focus to individual local objectives, often by incorporating meta-functions into the definition of the global learning objective~\cite{ Fallah2020PersonalizedFL}.

In student modeling, this statistical heterogeneity property manifests along several dimensions. 
% FDKT~\cite{Wu2021FederatedDK} addresses the heterogeneity arising from student training data being split across different schools.
FDKT~\cite{Wu2021FederatedDK} deals with the challenge of heterogeneity stemming from the division of student training data across multiple schools.
% ., where the schools are not willing to share data due to privacy concerns. 
% Applying the FL principle, local models are trained for each school, and updating between the global and local models is coordinated without transferring the sensitive data between schools. 
Applying the FL principle, individual school-specific models are trained, and communication between the global and local models is managed without sharing data across schools.
% While this setup is representative of some real-world educational settings, it ignores the bigger picture of heterogeneity across student subgroups such as race, gender, and other demographic variables. 
Although this setup mirrors certain real-world educational scenarios, it overlooks the broader context of diversity among student subgroups, including factors like race, gender, and demographic variables.
In AI fairness research, it has been noted that minority groups with limited data can be overlooked during model training, leading to biases and ethical concerns~\cite{Perez2020InvisibleWE, Ntoutsi2020BiasID}.
This happens because AI models are frequently trained on datasets that contain more samples from majority population groups, causing them to be better suited for these groups while being less effective for minority groups with limited representation. This issue has been identified in various AI application domains, such as medicine~\cite{MacNamee2002ThePO}, language modeling~\cite{Dixon2018MeasuringAM}, and image recognition~\cite{Kolla2022TheIO}.
As a result, biased models tend to be less accurate and unfair when applied to data from underrepresented groups. In the online education field, data from user interactions are similarly skewed according to the demographics of students who have historically taken these courses the most, naturally creating biases in AI-driven student models~\cite{Gardner2019EvaluatingTF, Kung2020InterpretableMD}. This bias has been observed in several applications, such as knowledge tracing~\cite{Barrett2024ImprovingMF} and graduation prediction~\cite{Anderson2019AssessingTF} tasks, where AI models produce lower-quality outputs for students from underrepresented backgrounds. For instance, students from lower-income support levels are often disadvantaged in these educational predictions, further exacerbating fairness issues in AI models~\cite{Barrett2024ImprovingMF,Anderson2019AssessingTF}. Such challenges highlight the importance of addressing bias and ethical concerns in the training data to promote fairness and accuracy for all groups.
\color{black}
% Specifically, AI models are typically trained on datasets that have more representation from majority groups, causing the model to perform better for these groups and less effectively for minority groups with limited data \cite{MacNamee2002ThePO, VanderWal2022TheBO, Dixon2018MeasuringAM, Kolla2022TheIO}. This can result in models that are less accurate or fair for underrepresented populations.
% In online learning environments, overlooking minority groups in model training can lead to personalized learning paths that fail to cater to the diverse needs of all students, resulting in educational disparities by not adequately supporting the unique challenges faced by underrepresented students~\cite{Gardner2019EvaluatingTF, Kung2020InterpretableMD, Anderson2019AssessingTF}.
\color{black}

% The aggregation scheme in FDKT weights the local models according to ``\emph{data quality},'' measured through fitting local psychometric models such as classical test theory and item response theory. 
The aggregation mechanism in FDKT assigns weights to local models based on their ``\emph{data quality},'' which is assessed by fitting local psychometric models like classical test theory and item response theory.
% This results in an implicit downweighting of underrepresented student subgroups due to a lack of data volume, which is not acceptable in our setting since the data from each group contains important information about how these students learn. 
This leads to an implicit devaluation of underrepresented student subgroups due to insufficient data volume. However, this is unacceptable in our setting as the data from each group holds crucial information about these students' learning processes.
We address this in our work by developing a Personalized FL methodology which accounts for heterogeneity across courses and demographics in a multi-tiered student modeling framework.

\subsection{Activity-Based Student Modeling}
Our student modeling approach will build upon two significant insights from previous research in educational data mining that can further mitigate biases.
These will come into play in Sec.~\ref{sec:tasks} as we formalize two downstream student modeling applications for case-studies of our methodology.

% The first observation is that patterns in student learning activity -- such as video-watching behavior, interactions in discussion forums, and clickstream logs (see Figure~\ref{fig:setting_heat}(a)) -- contain signals of student performance in online courses~\cite{Brinton2016MiningMC,Chu2021ClickBasedSP,chen2018behavioral}. 
% Since each student tends to generate a substantial number of activity measurements throughout a course, incorporating them may alleviate sparsity issues faced by prediction models trained on subgroups with fewer students.
The first insight is that student learning activity patterns, encompassing actions like video viewing, participation in discussion forums, and clickstream logs (as depicted in Figure~\ref{fig:setting_heat}(a)), contain signals of student achievement in online courses~\cite{Brinton2016MiningMC,Chu2021ClickBasedSP}. Given that each student typically generates a significant number of activity records over a course's duration, inclusion of such data could alleviate the sparsity challenges faced by prediction models trained on subgroups with fewer students.

The second observation is that unique learning activity patterns can be identified both within and across student groups (e.g., different gender groups)~\cite{Neill2019StructuredLE, Aguillon2020GenderDI}. 
% This was originally discovered among in-person learning behaviors (e.g., participation and learning style), and more recently in digital environments~\cite{McBroom2020HowDS}. 
This phenomenon was initially found in traditional classrooms, including behaviors like participation and {engagement with course materials}\color{black}, and more recently in digital learning environments as well~\cite{McBroom2020HowDS}.
% We thus aim to capture how sequences of learning activities (e.g., viewing a forum, then watching a specific video, then answering a specific quiz question) differ based on student subgroups in our construction of personalized FL models. 
Hence, we aim to consider how variations in sequences of learning behaviors, such as accessing a forum, viewing specific videos, and answering particular quiz questions, across student subgroups can be integrated into personalized federated learning models.
% However, online learning activities contain more noise than those captured in-person (e.g., a student accessing a video accidentally), making it difficult to identify these patterns through standard data mining techniques~\cite{Brinton2016MiningMC}. 
Yet, online learning behaviors are noisier than those occurring in in-person settings (e.g., accidental video access, inadvertent skipping through a video), posing challenges in identifying these patterns through conventional data mining methods~\cite{Brinton2016MiningMC}.

For example, considering one of the datasets in this paper, Figure~\ref{fig:setting_heat}(b) depicts the heatmaps of differences observed when students engage in particular learning activities.
% Two cases are considered: (b1) between learners across a demographic group (all males vs. all females), and (b2) between learners within a subgroup (50\% of the males vs. the other 50\%, chosen randomly). 
% Each heatmap value indicates the difference in the fraction of students engaged in the activity at that point in their learning trajectory. 
We explore two cases: (b1) across learners of a demographic group (all males vs. all females), and (b2) within a subgroup (50\% of males vs. the other 50\%, randomly chosen). Each heatmap value represents the difference in the proportion of students participating in the activity at that specific point in their learning journey.
% (b1) exhibits a more varied set of trajectories than (b2), consistent with the intuition that different subgroups have different learning behaviors, but (b2) still has noticeable differentials even though it is a comparison among a cross-section of the same subgroup.
(b1) reveals varied trajectories compared to (b2), consistent with diverse learning behaviors among different subgroups. However, (b2) also shows significant differences despite being a comparison within the same subgroup.
% This motivates the meta-learning approach we propose for subgroup personalization, where any machine-identified commonalities in the data across the subgroups are captured in a global model that is further refinable based on local subgroup information. 
This motivates the meta-learning approach we propose for subgroup personalization, wherein shared patterns in data across subgroups are integrated into a global model, which can then be fine-tuned using local subgroup information.
% As we will see in Sec.~\ref{sec:tsne}, visualizations of the student embeddings learned by our methodology validate that distinctive activity patterns exist between subgroups.
% \yun{for short}

\subsection{Outline and Summary of Contributions}
In this paper, we develop a federated student modeling methodology that personalizes prediction models for different data grouping criteria and mitigates the bias in data availability for underrepresented groups.
Specifically, we make the following major contributions:
\begin{itemize}
\item We design Multi-Layer Personalized Federated Learning (MLPFL), where local models associated with different student groupings are adapted from a global model (Section \ref{sec:method}). 
% This adaptation is in the form of meta-gradient updates taken on data that is localized to the specific course and/or student subgroup, instead of using data quality heuristics as in prior work, mitigating biases arising from heterogeneous data availability.
Unlike prior methods that invoke data quality heuristics, this adaptation involves meta-gradient updates on localized data from particular courses or student subgroups. We show that this mitigates biases from heterogeneous data availability.

\item We consider two popular downstream student modeling tasks for our methodology, knowledge tracing and outcome prediction (Section \ref{sec:down}). We formulate these based on multi-modal activity logs on learner interactions with course content and in discussion forums available in real-world online course datasets (Section \ref{sec:data-task}). 

\item Through experiments on three online education datasets (Section \ref{sec:exp}), we demonstrate that our methodology significantly outperforms existing approaches in terms of improving prediction accuracy and reducing variance across subgroups (Section \ref{sec:mdlbase}-\ref{sec:result}). We find this advantage appears at both the global and local model levels, resulting in a robust prediction framework that can adapt between different student groupings.  

\item Our experiments reveal that demographic-based MLPFL provides substantial improvements in student modeling compared to course-level personalization without demographic information. 
% We also visually demonstrate that the student activity embeddings produced by our method are well organized based on subgroups (Section \ref{sec:tsne}). 
We also visually examine how the student activity embeddings generated by our approach cluster according to subgroups, and find that they exhibit informative clusterings (Section \ref{sec:tsne}).

%  \yun{(1) we show the amount of how hyrchacal info provided affect the personalization and (2) tsne visuals show impact that by subgroup glean better info when detailed info is provided.}
% \yun{course and subgroup, amount of how hyrchacal info provided}
% \yun{glean better info (subgroup)}

\end{itemize}

This paper is an extension of our prior conference version~\cite{https://doi.org/10.48550/arxiv.2208.01182}. 
Compared with~\cite{https://doi.org/10.48550/arxiv.2208.01182}, this extension adds the following major components: 
% We consider learning patterns across different courses, rather than just within courses. 
% We enhance personalized federated learning by incorporating structural and layer-wise aspects, exploring how varying hierarchies impact training and personalization outcomes. 
% We introduce subgroup-level, course-level and global-level hierarchies in this extension and our MLPFL method can adeptly capture meaningful information within each hierarchy.
% We extend our exploration to include additional prediction tasks and consider scenarios with and without student information. 
% Additionally, our analysis involves various baselines, enabling a systematic comparison of personalization effects across hierarchies.
(1) We consider a \emph{hierarchical} personalization approach that incorporates subgroup variables across both courses and demographic groups within courses, while ~\cite{https://doi.org/10.48550/arxiv.2208.01182} only considers demographic groups and treats each course independently.
(2) In addition to student outcome prediction considered in~\cite{https://doi.org/10.48550/arxiv.2208.01182}, we extend our exploration to include the knowledge tracing use case. 
(3) Our experiments have been augmented with various baselines to evaluate different components of hierarchical personalization, and have also introduced a practical scenario where access to students' demographic information is unavailable.
% Overall, our research showcases how FL innovations can enhance personalized student modeling and provide insights into the importance of fair and unbiased learning algorithms. 

\section{Multi-Layer Personalized Federated Learning}
\label{sec:method}
\begin{figure*}[t]
    \centering
    \setlength{\abovecaptionskip}{1mm}
    \includegraphics[width=0.85\linewidth]{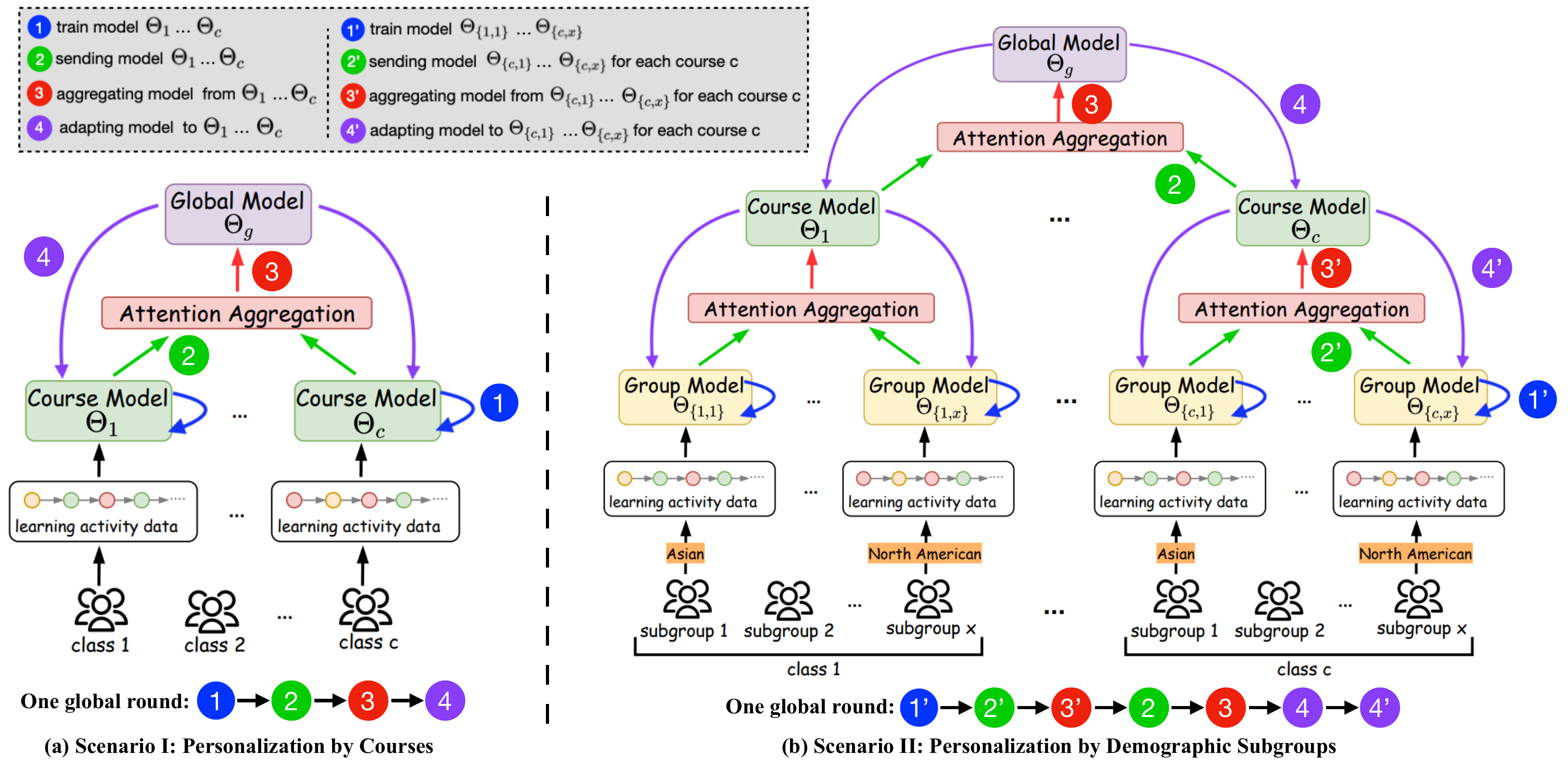}
	 \caption{Overview of the personalized methodology we develop in this paper for customizing prediction models for (a) each course and (b) each demographic student subgroup. 
  Our Multi-Layer Personalized Federated Learning (MLPFL) methodology refines global models in a hierarchical manner to create personalized local student models for distinct student groups, leveraging both commonality and subgroup diversity.
  }
\label{fig:overview}
\vspace{-0.2in}
\end{figure*}

This section introduces the training procedures of Multi-Layer Personalized Federated Learning (MLPFL) based on general modeling assumptions. We detail the model structures for specific downstream tasks in Section \ref{sec:down}.
The two layers we consider for our modeling hierarchy are \textit{courses} and \textit{demographics}. Since demographic information may not always be available, we consider two different scenarios, as depicted in Figure \ref{fig:overview}.
The first scenario (Figure \ref{fig:overview}(a)) conducts adaptable student modeling by course only.
The second scenario (Figure \ref{fig:overview} (b)) further considers personalization for each demographic subgroup within each course, i.e., when students have provided this information. 

\vspace{-0.15in}

\subsection{Scenario I:  Course-specific Adaptation \color{black}}
\label{sec:sc1_course}
\subsubsection{Data Partitioning by Courses}
\label{sec:sc1_data}
Well-defined dataset partitioning is essential for scenarios I and II since it is the basis for personalization.
We use $\Omega_c$ to represent the students of each course $c \in \{1, 2, ..., C\}$, where $C$ is the number of courses.
For each course, the partition is done by splitting students into train-test sets at a ratio of 4:1, denoted as $\Omega_c = \{\Omega^{\mathsf{train}}_c, \Omega^{\mathsf{test}}_c\}$.
Therefore, our entire collected dataset is $\Omega = \bigcup_{c=1}^{C} \Omega_c = \{\Omega^{\mathsf{train}}, \Omega^{\mathsf{test}}\}$, where $\Omega^{\mathsf{train}} = \bigcup_{c=1}^{C} \Omega^{\mathsf{train}}_c$ and $\Omega^{\mathsf{test}} = \bigcup_{c=1}^{C} \Omega^{\mathsf{test}}_c$.
The methodology proposed in this section uses $\Omega^{\mathsf{train}}_c$ ($c=1, \cdots, C$) to build $C$ models for courses, denoted as $\Theta_c$, for the downstream tasks formulated in Section~\ref{sec:tasks}. 
The prediction performance of each trained model $\Theta_c$ is subsequently evaluated by using test set $\Omega^{\mathsf{test}}_c$.

\subsubsection{Meta-Learning Adaptation}
\label{sec:person_course}
To facilitate adaptation to courses\color{black}, our methodology obtains separate local models for each course. 
% The tasks handled locally are: (i) providing a course-personalized model $\Theta_c$ for global aggregations and (ii) adapting global model  $\Theta_g$  based on the students in each course. 
% We aim to train a global model that is \textit{easily adaptable} to each course. 
Locally, we address two tasks: (i) generating a course-customized model $\Theta_c$ for global aggregations and (ii) adapting the global model $\Theta_g$ according to the characteristics of students within each course. Our objective is to develop a global model that can be \textit{easily adaptable} to each course.
% To this end, we employ a meta-learning based personalized FL framework~\cite{Fallah2020PersonalizedFL} to seek the global model that solves the optimization problem:
For this purpose, we employ a personalized federated learning framework based on meta-learning~\cite{Fallah2020PersonalizedFL} to find the global model that addresses the optimization problem:
\begin{equation}
\label{eq:mainLoss}
\operatorname \min_{\Theta} \sum_{c\in\mathcal{C}} F_c(\Theta) := f_c \big(\underbrace{\Theta - \nabla f_c(\Theta)}_{(a)}\big),
\end{equation}
where $F_c(\cdot)$ is the meta-loss function of course $c$, and $f_c(\cdot)$ is the original loss function, which is the sum of prediction losses (i.e., binary cross entropy loss as we will formulate in Section \ref{sec:kt} and \ref{sec:outcomeprediction}) over course $c$.
% The main difference between our loss function in (\ref{eq:mainLoss}) and the loss functions used in existing student modeling work~\cite{dkt,botelho2017improving} is that our loss function aims to minimize the loss of the \emph{adapted} versions of the global model.
The key distinction between our loss function (equation (\ref{eq:mainLoss})) and those used in prior student modeling studies~\cite{dkt,botelho2017improving} is that ours minimizes the loss of the \emph{adapted} versions  of the global model.
% This adaptation is based on one gradient descent step, i.e., term $(a)$ in (\ref{eq:mainLoss}), which is taken over the dataset of course $c$. That is, our method exploits the \textit{commonality} of data across courses to train an adaptable global model, which can be easily tailored to each course.
This adaptation is performed through a single gradient descent step, represented as term $(a)$ in (\ref{eq:mainLoss}), executed on the dataset of course $c$. In essence, our approach leverages the \textit{commonality} of data across courses to develop an adaptable global model which can be easily tailored to individual courses.

% To solve (\ref{eq:mainLoss}), we derive meta-gradient based local update steps. In particular, training proceeds through a sequence of training \textit{rounds} $k\in\{1,\cdots,K\}$, with each round consisting of multiple local training \textit{iterations} $e\in\{1,\cdots,E\}$.
% In each iteration $e$, course $c\in \mathcal{C}$ updates its local model $\Theta_c^{(k,e)}$ using only the students data from this course. 
To address (\ref{eq:mainLoss}), we derive local update steps based on meta-gradients. The training process is executed in a series of training \textit{rounds} $k\in{1,\cdots,K}$, where each round comprises multiple local training \textit{iterations} $e\in{1,\cdots,E}$. Within each iteration $e$, course $c\in \mathcal{C}$ updates its local model $\Theta_c^{(k,e)}$ using solely the student data associated with that course.
% These local models are synchronized through a global aggregation step at the end of each round.
After each local model completes its $E$ iterations\color{black}, these local models are synchronized through a global aggregation process.
% Specifically, in each round $k$, the local model is initialized as $\Theta_c^{(k,0)}=\Theta_{{g}}^{(k)}$, $\forall c$, where $\Theta_{{g}}^{(k)}$ is the global model obtained at the end of round $k-1$. Then, each course $c$ conducts its meta-gradient updates according to
In every round $k$, the local model for each course $c$ begins with initialization as $\Theta_c^{(k,0)}=\Theta_{{g}}^{(k)}$, where $\Theta_{{g}}^{(k)}$ represents the global model achieved at the completion of round $k-1$. Then, each course $c$ performs its meta-gradient updates according to
\begin{equation}\label{eq:metaUpd}
    \Theta_c^{(k,e)}=\Theta_c^{(k,e-1)}-\eta \nabla F_c \left(\Theta_c^{(k,e-1)}\right),~e=1,\cdots,E,
\end{equation}
where $\eta$ is the step size, and following (\ref{eq:mainLoss}), the meta gradient $\nabla F_c$ is computed as:
\begin{equation} \label{eq:metaGradd}
\nabla F_c(\Theta) = \left(\mathbf{I} - \nabla^2 f_c(\Theta)\right)\nabla f_c\left(\Theta - \nabla f_c(\Theta)\right),~\forall \Theta,\hspace{-1mm}
\end{equation}
where $\nabla^2$ is the Hessian operator. 
% The second-order Hessian term in (\ref{eq:metaGradd}) can be well-approximated by first-order methods without incurring significant performance degradation~\cite{Fallah2020PersonalizedFL}. 
The second-order Hessian term in (\ref{eq:metaGradd}) can be first-order approximated without causing deterioration in performance~\cite{Fallah2020PersonalizedFL}.
% After local updates, the global aggregation $\Theta_{{g}}^{(k+1)}$ is obtained using an attention-based method that we detail next.
Following the local updates, we compute the global aggregation $\Theta_{{g}}^{(k+1)}$ through an attention-based technique explained in the following section.

\subsubsection{Attention-Based Global Model Aggregation}
\label{sec:scen1_agg}
% The global modeling stage in our methodology handles two tasks: (i) aggregating the local models and (ii) synchronizing the course models with the resulting parameter vector for subsequent local model training rounds.
The global modeling phase in our approach involves two tasks: (i) aggregating local models and (ii) synchronizing course models with the resulting parameter vector for subsequent rounds of local model training.
% Instead of using the standard averaging-based aggregation method from FedAvg~\cite{McMahan2017CommunicationEfficientLO}, we employ a recently developed attention-based aggregation method, FedAtt~\cite{Ji2019LearningPN}.
Rather than relying on the conventional averaging-based aggregation FedAvg~\cite{McMahan2017CommunicationEfficientLO}, we utilize an attention-based aggregation, FedAtt~\cite{Ji2019LearningPN}.
% FedAtt introduces an attention mechanism for weighting local neural network models according to how far the parameters in each of their layers have deviated from the previous global model.
FedAtt incorporates an attention mechanism to assign weights to local models based on the extent to which the parameters in their individual layers have diverged from the previous global model.

% Formally, at each aggregation step, the global model receives two pieces of information: (i) all the local models $\Theta^{(k,E)}_{c}$ ($c=1, \cdots, C$), and (ii) the attention weight for each local model.
In each aggregation step, the global model is informed by two elements: (i) the entirety of local models $\Theta^{(k,E)}_{c}$ ($c=1, \cdots, C$), and (ii) the respective attention weight associated with each local model.
% The attention weight $\alpha^{(k)}_c$ for local course model $\Theta_c$ is computed across its set of layers $\mathcal{L}$ as follows:
The attention weight $\alpha^{(k)}_c$ for the local course model $\Theta_c$ is calculated across its layers $\mathcal{L}$ as follows:
\begin{equation}\label{eq:globalatt}
\alpha^{(k)}_c = \sum_{\ell \in\mathcal{L}} \textup{softmax}\left(\left\Vert\Theta_{{g}}^{(k)}({\ell}) - \Theta_c^{(k,E)}(\ell)\right\Vert\right),
\end{equation}
% where $\ell \in \mathcal{L}$ denotes a particular neural network model layer, and $\Theta^{(k)}_{g}(\ell)$ and $\Theta^{(k,E)}_{c}(\ell)$ represent the parameter vector of the $\ell$-th layer in the global and local model, respectively, at the instance of global aggregation.
 where $\ell \in \mathcal{L}$ represents a specific model layer, and $\Theta^{(k)}_{g}(\ell)$ and $\Theta^{(k,E)}_{c}(\ell)$ denote the parameter vectors of the $\ell$-th layer in the global and local models, respectively, during global aggregation.
The resulting aggregated global model is:
\begin{equation}\label{eq:globalupdate}
\Theta_{{g}}^{(k+1)} = \Theta_{{g}}^{(k)} - \epsilon \sum_{c\in\mathcal{C}} \alpha^{(k)}_c\left(\Theta_{{g}}^{(k)} - \Theta_{c}^{(k,E)}\right),
\end{equation}
where $\epsilon$ is a tunable step-size.
The full personalized FL process is outlined in Algorithm~\ref{tb:server_client_sc1} and Figure~\ref{fig:overview}(a).

\begin{algorithm}[t]
\algtext*{EndFor}

\caption{Course Level MLPFL}
% \caption{Course Level HPFL}

\begin{algorithmic}[1]
\State Global Model at $k$ training round: $\Theta_g^k$; Local Model of course $c$ at $k$ training round and $e$ local iteration: $\Theta_c^{(k,e)}$

\State \textbf{Global Execution}:
\State Initialize global model $\Theta_\mathsf{g}^{(0)}$
\For{each global round $k = 1, 2,..., K$}
\For{ each course $c\in\mathcal{C}$ \textbf{in parallel}}
\State $\Theta_c^{(k,E)} \gets$ \textbf{LocalAdaptation ($\Theta_{\mathsf{g}}^{(k-1)}$\color{black})}
\EndFor
\For{ each course $c\in\mathcal{C}$}
% \For{ each layer $\ell \in \mathcal{L}$}
\State Compute attention weight $\alpha_c^{(k-1)}$ \color{black} using (\ref{eq:globalatt})
% \State $s_x(\ell) = ||\Theta_g(\ell) - \Theta_x(\ell)||$
% \State $\alpha_x(\ell) = \textup{softmax}(s_x(\ell))$
% \EndFor
% \State $\alpha_x = \sum_{l} \alpha_x(\ell)$
\EndFor
\State Obtain the global model $\Theta_{\mathsf{g}}^{(k)}$ \color{black} by (\ref{eq:globalupdate})
% $\Theta_g^{(k+1)} \gets \Theta_g^{(k)} - \epsilon \sum_{x \in\mathcal{X}} \alpha_x(\Theta_g^{(k)} - \Theta_x^{(k+1)})$
\EndFor
\end{algorithmic}

\begin{algorithmic}[1]
\State \textbf{LocalAdaptation ($\Theta^{(k-1)}_g$\color{black})}:
\State Initialize the local model $\Theta_c^{(k,0)} = \Theta_\mathsf{g}^{(k-1)}$  \color{black} 
\For {Each local iteration $e=1,\cdots,E$}
\State Obtain $\Theta_c^{(k,e)}$ using the meta-update rule (\ref{eq:metaUpd})
% \For{batch $b \in B$}
% \State $\widetilde{\Theta} \gets \Theta - \eta\nabla\ell(\Theta;b)$
% \EndFor
% \State $\Theta \gets \Theta - \zeta\nabla \sum \ell(\widetilde{\Theta})$

\EndFor
\State Return parameters $\Theta^{(k,E)}_c$

\end{algorithmic}
\vspace{-0.05in}

\label{tb:server_client_sc1}
\end{algorithm}

\subsection{Scenario II: Personalized by Course and Demographics}
\label{sec:sc2_demo}
\subsubsection{Data Partitioning by Demographic Variables}
Open platforms, such as edX and Coursera, usually solicit students' demographic details during registration, which they can fill out voluntarily.
In this section, we add an additional layer of personalization based on this demographic information.
Next, we illustrate the partitioning criterion using the training set as an example; the test set follows the same partitioning rule.
We use $\Omega_{\{c, I\}}^{\mathsf{train}}$ to represent the students belonging to the training set who answered specific demographic information $I$ within course $c$. 
For each variable $I$, we assume there are a finite set of categories (groups) for students to select from, e.g., as defined by a dropdown list. 
Based on our available datasets from edX (see Section~\ref{sec:data-task}), the demographic information $I \in \{\texttt{G, C, Y}\}$ includes three frequently employed variables: Gender (\texttt{G}), Country (\texttt{C}), and Year of Birth (\texttt{Y}).
For country, we group students into five continents since some countries receive no responses at all.
$\Omega_{\{c,\texttt{G}\}}^{\mathsf{train}}$ , $\Omega_{\{c,\texttt{C}\}}^{\mathsf{train}}$, and $\Omega_{\{c,\texttt{Y}\}}^{\mathsf{train}}$ are not necessarily the same since students may choose to not respond to certain demographic questions. 

For the set $\Omega_{\{c, I\}}^{\mathsf{train}}$, we further define $\Omega_{\{c, I, x\}}^{\mathsf{train}}$ as the set of students belonging to subgroup $x \in \mathcal{X}$, where $\mathcal{X}$ is the set of groups in variable $I$. 
For example, $\mathcal{X} = \{\texttt{M} ,\texttt{F}\}$ represents male and female subgroups for gender information ($I =$ \texttt{G}) when it is provided. 
For country information ($I = $ \texttt{C}), $\mathcal{X} = \{\texttt{AS}, \texttt{AF}, \texttt{EU}, \texttt{NA}, \texttt{SA}\}$ represents Asian (\texttt{AS}), African (\texttt{AF}), European (\texttt{EU}), North American (\texttt{NA}), and South American (\texttt{SA}) student subgroups. 
Furthermore, $\mathcal{X} = \{\texttt{<80}, \texttt{80-90}, \texttt{>90}\}$ represent year of birth prior to 1980, between 1980 and 1990, and after 1990 for birth year information ($I = Y$).
To conclude, $\Omega_{\{c, I\}}^{\mathsf{train}} = \bigcup_{x \in \mathcal{X}} \Omega_{\{c, I, x\}}^{\mathsf{train}}$

Moreover, we introduce $\Omega_{\{c, \neg I\}}^{\mathsf{train}}$ to denote the set of students that chose not to provide demographic information $I$ in the training set of course $c$.
To be specific, $\Omega_c^{\mathsf{train}} = \{\Omega_{\{c, I\}}^{\mathsf{train}}, \Omega_{\{c,\neg I\}}^{\mathsf{train}}\}$.
We will use $\Omega_c^{\mathsf{train}}$ to build several local models, denoted as $\Theta_{\{c, I, x\}}$ or $\Theta_{\{c, \neg I\}}$.
$\Theta_{\{c, \neg I\}}$ denotes the model built by $\Omega_{\{c, \neg I\}}^{\mathsf{train}}$, and $\Theta_{\{c, I, x\}}$ represents the model constructed by the specific $x$ subgroup of demographic information $I$ within course $c$. 
When performing Multi-Layer Personalized Federated Learning in this section, we consider two settings: (i) training models by using $\Theta_{\{c, I, x\}}$ and (ii) training an additional model $\Theta_{\{c, \neg I\}}$ alongside $\Theta_{\{c, I, x\}}$.
The prediction performance of each model then is tested on $\Omega_{\{c,I,x\}}^{\mathsf{test}}$ or $\Omega_{\{c,\neg I\}}^{\mathsf{test}}$ after the training procedure.

\subsubsection{Meta-Learning Personalization}
\label{sec:scen2_meta}
We build separate models for demographic subgroups to facilitate more fine-granular prediction personalization within each course.
Specifically, as shown in Figure~\ref{fig:overview}(b),  we build separate subgroup-personalized models and increase the hierarchy during MLPFL training.
% Similar to the meta-learning local model personalization for courses in Sec.~\ref{sec:person_course}, the tasks handled locally are: (i) providing a subgroup-personalized model for global aggregations and (ii) adapting the global model based on the local subgroup dataset. 
Similar to the local model personalization for courses as explained in Section~\ref{sec:person_course}, we undertake two tasks at the subgroup level: (i) generating a subgroup-personalized model for global aggregations and (ii) adapting the global model using local subgroup data.
Our aim is to train a global model that is \textit{easily adaptable} to each local student subgroup. 
Therefore the optimization problem will change from (\ref{eq:mainLoss}) to:
\begin{equation}
\label{eq:demoLoss}
\operatorname \min_{\Theta} \sum_{c\in\mathcal{C}} \sum_{x\in\mathcal{X}} F_{\{c,I,x\}}(\Theta) := f_{\{c,I,x\}} \big(\underbrace{\Theta - \nabla f_{\{c,I,x\}}(\Theta)}_{(b)}\big),
\end{equation}
where $F_{\{c,I,x\}}(\cdot)$ is the meta-loss function of student subgroup $x$ (corresponding to information $I$) in course $c$, and $f_{\{c,I,x\}}(\cdot)$ is the original loss function, which is the sum of prediction loss over the dataset of group $x$ in course $c$. 
% The adaptation is based on one gradient descent step, i.e., term $(b)$ in (\ref{eq:demoLoss}), which is taken over the local dataset of subgroup $x$. That is, our method exploits the \textit{commonality} of data across subgroups to train an adaptable global model, which can be easily tailored to each individual subgroup.
This adaptation is performed through a single gradient descent step, indicated by term $(b)$ in (\ref{eq:demoLoss}), conducted on the local dataset of subgroup $x$. In essence, our approach leverages the \textit{commonality} of data across subgroups to develop an adaptable global model, readily customizable for each specific subgroup.

To solve (\ref{eq:demoLoss}), we derive meta-gradient based local update steps. 
Similar to Scenario I, the training also proceeds through a sequence of training rounds $k\in\{1,\cdots,K\}$, with each round consisting of multiple local training iterations $e\in\{1,\cdots,E\}$.
During each iteration $e$, the local subgroup-model $\Theta_{\{c,I,x\}}^{(k,e)}$ is updated by meta-gradient:
\begin{equation}\label{eq:metaUpd_demo}
    \Theta_{\{c,I,x\}}^{(k,e)}=\Theta_{\{c,I,x\}}^{(k,e-1)}-\eta \nabla F_{\{c,I,x\}} \left(\Theta_{\{c,I,x\}}^{(k,e-1)}\right),~e=1,\cdots,E,
\end{equation}
where $\eta$ is the step size, and $\nabla F_{\{c,I,x\}}$ is the meta gradient.

\subsubsection{Global Model Aggregation and Adaptation}
\label{sec:scen2_agg}
We employ attention-based aggregation for global model construction.
As we have additional hierarchy for training, the global modeling stage in this scenario encompasses four tasks: (i) aggregating local subgroup-models into course-models, (ii) aggregating course-models into a global model, (iii) synchronizing course-models with the resulting global parameters for course-level adaptation, and (iv) synchronizing subgroup-models with course-level parameters for subgroup-level adaptation.

At each aggregation step, local subgroup-models are first aggregated into course-models as:
\begin{equation}\label{eq:course_agg}
\Theta_{{c}}^{(k+1)} = \Theta_{c}^{(k)} - \epsilon \sum_{x\in\mathcal{X}} \alpha^{(k)}_{\{c,I,x\}}\left(\Theta_{c}^{(k)} - \Theta_{\{c,I,x\}}^{(k,E)}\right) \forall I,
\end{equation}
where $\alpha^{(k)}_{\{c,I,x\}}$ is the layer-wisely computed attention weight for local subgroup-models:
\begin{equation}\label{eq:course_agg_att}
\alpha^{(k)}_{\{c,I,x\}} = \sum_{\ell \in\mathcal{L}} \textup{softmax}\left(\left\Vert\Theta_{{c}}^{(k)}({\ell}) - \Theta_{\{c,I,x\}}^{(k,E)}(\ell)\right\Vert\right),
\end{equation}
with $\mathcal{L}$ denoting the set of layers.
After aggregating the subgroup-models into course-models, the global model $\Theta_g$ receives the parameters of each course-model $\Theta_c$ and performs attention-based aggregation as follows:
\begin{equation}\label{eq:global_agg}
\Theta_{\mathsf{g}}^{(k+1)} = \Theta_{\mathsf{g}}^{(k)} - \epsilon \sum_{c\in\mathcal{C}} \alpha^{(k)}_c\left(\Theta_{\mathsf{g}}^{(k)} - \Theta_{c}^{(k+1)}\right),
\end{equation}
where $\alpha^{(k)}_c$ is the attention weight for course-models:
\begin{equation}\label{eq:global_agg_att}
\alpha^{(k)}_c = \sum_{\ell \in\mathcal{L}} \textup{softmax}\left(\left\Vert\Theta_{\mathsf{g}}^{(k)}({\ell}) - \Theta_c^{(k+1)}(\ell)\right\Vert\right),
\end{equation}
and $\ell \in \mathcal{L}$ denotes a particular model layer.

The full MLPFL algorithm is summarized in Algorithm~\ref{tb:server_client_sc2} and Figure~\ref{fig:overview}(b). Note that, in Scenario II, there are two steps of synchronization after $\Theta_{\mathsf{g}}^{(k)}$ has been aggregated, compared with Scenario I where there is just one step. The first synchronization is adapting the global model $\Theta_g$ as a temporary course-model $\Theta_c'$, as shown in line 6 of Algorithm~\ref{tb:server_client_sc2}. This course adaptation is done through a single-step update by using a batch of students from course $c$. We use stratified sampling to select a few students within each subgroup to form this batch without bias. After the course adaptation, the second step is to synchronize and adapt the subgroup models, shown in line 8. In each round $k$, the subgroup model for demographic $x$ in course $c$ is initialized as $\Theta_{c,x}^{(k,0)}=\Theta_{{c}}^{(k)'}$, where $\Theta_{{c}}^{(k)'}$ is the temporary course-model. 
Referring to Figure~\ref{fig:overview}(b), the $\Theta_{c,x}^{(k,E)}$ correspond to the models at the bottom level of the hierarchy, which have the most fine-granular adaptation. Then, in line 11, the updated course-level models $\Theta_{{c}}^{(k+1)}$ are obtained by aggregating the bottom-level models, corresponding to the middle level of the hierarchy. Finally, in line 14, the updated global model $\Theta_{\mathsf{g}}^{(k+1)}$  is obtained by aggregating the middle-level models, corresponding to the top level of the hierarchy.
\color{black}

\begin{algorithm}[t]
\algtext*{EndFor}

\caption{Demographic Level MLPFL}
\begin{algorithmic}[1]
\State Global Model at $k$ training round: $\Theta_g^k$; Global Course-level Model of course $c$ at $k$ training round: $\Theta_c^{k}$; Local Demographic-level Model of demographic subgroup $x$ in course $c$ at $k$ training round and $e$ local iteration: $\Theta_{\{c,x\}}^{(k,e)}$

\State \textbf{Global Execution}:
\State Initialize global model $\Theta_\mathsf{g}^{(0)}$

\For{each global round $k = 1, 2,..., K$}
\For{each course $c\in\mathcal{C}$ \textbf{in parallel}}
\State  $\Theta_{c}^{'(k)}$ \color{black} $\gets$ \textbf{Course-level Adaptation ($\Theta_{\mathsf{g}}^{(k-1)}$\color{black})}
\EndFor

\For{each subgroup $x\in\mathcal{X}$ \textbf{in parallel}}
\State $\Theta_{\{c,x\}}^{(k,E)} \gets$ \textbf{LocalAdaptation ($\Theta_{c}^{'(k)}$\color{black}) }
\EndFor
\For{each subgroup $x\in\mathcal{X}$}
\State Compute attention weight $\alpha_x^{(k-1)}$ \color{black} using (\ref{eq:course_agg_att})
\EndFor
\State Obtain the course-level model $\Theta_{\mathsf{c}}^{(k)}$ \color{black} by (\ref{eq:course_agg})

\For{ each course $c\in\mathcal{C}$}
\State Compute attention weight $\alpha_c^{(k-1)}$ \color{black} by (\ref{eq:global_agg_att})

\EndFor

\State Obtain global model $\Theta_{\mathsf{g}}^{(k)}$ \color{black} based on (\ref{eq:global_agg})
% $\Theta_g^{(k+1)} \gets \Theta_g^{(k)} - \epsilon \sum_{x \in\mathcal{X}} \alpha_x(\Theta_g^{(k)} - \Theta_x^{(k+1)})$
\EndFor
\end{algorithmic}

\vspace{0.05in}

\begin{algorithmic}[1]
\State \textbf{Course-level Adaptation ($\Theta^{(k-1)}_g$\color{black})}:
\State Initialize the course-level model $\Theta_{c}^{(k)} = $ $\Theta^{(k-1)}_g$\color{black}
\State Obtain $\Theta_c^{'(k)}$ using one step of meta-update (\ref{eq:metaUpd})
\end{algorithmic}

\vspace{0.05in}

\begin{algorithmic}[1]
\State \textbf{LocalAdaptation ($\Theta_c^{'(k)}$)}:
\State Initialize the local model $\Theta_{\{c,x\}}^{(k,0)} = \Theta_c^{'(k)}$
\For {each local iteration $e=1,\cdots,E$}
\State Obtain $\Theta_c^{(k,e)}$ using the meta-update (\ref{eq:metaUpd_demo})

\EndFor
\State Return parameters $\Theta^{(k,E)}_c$ for each course $c$
\end{algorithmic}

\label{tb:server_client_sc2}
\end{algorithm}

\section{Datasets and Prediction Tasks}
\label{sec:tasks}

% In this section, we describe the datasets we consider and formalize the downstream student modeling prediction tasks we employ to evaluate our methodology.

\subsection{Datasets}
\label{sec:data-task}
% \yun{combine with prediction tasks}
% In this section, we will introduce the dataset and the prediction task, including prediction objective, data encoding, and model structure.
% We collect our datasets from Purdue University's online courses on the edX platform (\url{www.edx.org}). We study three graduate-level courses: Fiber Optic Communications (FOC), Quantum Detectors and Sensors (QDS), and Essentials of MOSFETs (MOSFETs).
Our datasets are sourced from online courses hosted on the edX platform (\url{www.edx.org}) at Purdue University.
We study three graduate-level courses: Fiber Optic Communications (FOC), Quantum Detectors and Sensors (QDS), and Essentials of MOSFETs (MOSFETs).
% Each course contains a series of lecture videos, some of which have an end-of-video quiz that assesses student learning progress.
% Each course also provides a discussion forum page for students to interact with each other.
Within each course, a series of lecture videos is available, some of which include end-of-video quizzes to evaluate student comprehension. Additionally, every course features a discussion forum where students can engage in interactions with one another.
% The platform also records each student's final grade, which is a pass or fail label based on the grading policy.
The platform also logs the final grade of each student, indicating whether they passed or failed according to the grading policy.
% Summary statistics of the three courses are given in Table~\ref{tb:course}, emphasizing a broad range of activity levels for our evaluation.
Table~\ref{tb:course} provides summary statistics for the three courses, highlighting a diverse range of activity levels for our evaluation.
Details about the dataset are available in Appendix~\ref{sec:data-dist}.
% \begin{table}
% \caption{Summary details of three datasets acquired from edX.}

% \begin{center}
% \scalebox{0.9}{
% \begin{tabular}{lcccc}
% \hline
% & FOC & QDS & MOSFETs\\
% \hline\hline
% \# of students & 1,265 & 2,304 & 886\\
% \# of lecture videos & 43 & 31 & 26\\
% \# end-of-video quizzes & 25 & 23 & 11\\
% \# of discussion threads & 20 & 35 & 17\\
% Avg. reply per thread & 1.95 & 0.48 & 1.44\\
% \# of activities & 63,789 & 95,487 & 31,382\\
% video activity vs. forum activity & 67\%, 33\% & 70\%, 30\% & 59\%, 41\%\\
% final pass rate & 29.3\% &	24.8\% & 31.1\%\\
% \hline
% \end{tabular}}
% \end{center}
% % \vspace{-3mm}
% \vspace{-0.2in}
% \label{tb:course}
% \end{table}

\begin{table}
\caption{Summary details of three datasets acquired from edX.}

\begin{center}
\scalebox{0.9}{
\begin{tabular}{lcccc}
\hline
& FOC & QDS & MOSFETs\\
\hline\hline
\# of students & 1,265 & 2,304 & 886\\
\# of lecture videos & 43 & 31 & 26\\
\# end-of-video quizzes & 25 & 23 & 11\\
\# of discussion threads & 20 & 35 & 17\\
Avg. reply per thread & 1.95 & 0.48 & 1.44\\
Avg. activities per student & 50.42 & 41.44 & 35.41\\
% video activity vs. forum activity & 67\%, 33\% & 70\%, 30\% & 59\%, 41\%\\
% final pass rate & 29.3\% &	24.8\% & 31.1\%\\
\hline
\end{tabular}}
\end{center}
% \vspace{-3mm}
\vspace{-0.3in}
\label{tb:course}
\end{table}
% \input{tables/combined_table}
% \vspace{-0.1in}

\subsubsection{Video-watching behavior and quiz responses}
Each time a student $u$ accesses a lecture video $v_c$ of course $c$, their activity is recorded with the following information: student ID, course ID, video ID, and UNIX timestamp.
% The in-video quizzes for each of the courses consist of a single multiple-choice question or a True/False question, appearing at the end of the video.
In each course, the in-video quizzes comprise either a single multiple-choice question or a True/False question, presented at the end of the video.
When student $u$ submits an answer to the in-video question for video $v_c$, the platform records the student's response $r_{u,v_c} \in \{0, 1\}$. 
$r_{u,v_c} = 0$ indicates an incorrect response, and $r_{u,v_c} = 1$ otherwise.

\subsubsection{Discussion forum interactions}
%Discussion forums create social networks among askers, answerers, and other students~\cite{Brinton2018OnTE, 6814139, Vieira2022StudyOT, Sahay2023PredictingLI, Liu2021ExploringTR}. In these forums, students learn from one another through structured interactions, making them an important part of student knowledge modeling.
% The learning platform records forum-participation activities $f_u \in \{\textsf{forum\_post}, \textsf{forum\_reply}, \textsf{forum\_view}\}$ when student $u$ visits the discussion forum page.
When student $u$ visits the discussion forum page, the learning platform logs forum participation activities as $f_u \in \{\textsf{forum\_post}, \textsf{forum\_reply}, \textsf{forum\_view}\}$.
% \textsf{forum\_post} indicates a student either started a new thread or made a post in a thread, \textsf{forum\_reply} represents a student replied to another student's post, and \textsf{forum\_view} indicates a student visited a thread without posting or replying.
\textsf{forum\_post} represents a student initiating a new thread or making a post to an existing one, \textsf{forum\_reply} indicates a student's response to another student's post, and \textsf{forum\_view} marks a student's visit to a thread without posting or replying.
\vspace{-0.1in}

\subsection{Student Prediction Modeling Tasks}
\label{sec:down}
To evaluate MLPFL, we implement two downstream tasks, knowledge tracing~\cite{dkt} and student outcome prediction~\cite{https://doi.org/10.48550/arxiv.2208.01182}.

\subsubsection{Knowledge Tracing}
\label{sec:kt}
We formulate knowledge tracing task by using students' video-and-quiz interactions in our dataset. Let
$x_{u,t} = (i_{u,t}, r_{u,v_c,t})$ be a tuple representing the item $i$ attempted by student $u$ at time $t$, where $t$ serves as the activity index for each student. $r_{u,v_c,t} \in \{0,1\}$ represents the result of the student's response to the quiz on video $v_c$ at time $t$.  $i_{u,t} = \mathbbm{1}(c) \oplus \mathbbm{1}(v_c)$ is a combined representation of course ID $c$ and video $v_c$ by using one-hot function $\mathbbm{1}(\cdot)$, where $\oplus$ denotes vector concatenation.

Given a series of interactions $X_u = \{x_{u,1}, x_{u,2}, ..., x_{u,t}\}$, the goal of knowledge tracing task is to predict the value of $r_{u,v_c,t+1}$ representing if student $u$ will answer the new item correctly based on their current knowledge state $\textbf{h}_{u,t}$.
The current knowledge state of a student will be highly correlated with their previous knowledge state. 
Therefore, we follow \cite{dkt}'s suggestion that models student learning process by Long Short Term Memory network (LSTM) as:
\begin{equation}
\textbf{h}_{u,t} = \textup{LSTM}\left(i_{u,t}, \; \textbf{h}_{u,t-1}\right).
\end{equation}
The probability of correctly answering item $i_{u,t}$ is:
% by student $u$:
\begin{equation} 
    \textbf{p}_{u,t+1} = \textup{softmax}\left(\textbf{W} \cdot {\textbf{h}}_{u,t} + \textbf{b}\right),\\
\end{equation}
where $\textbf{W}\in \mathbb{R}^{k \times 2}$ is the weight for linearly transforming ${\textbf{h}}_{u,t}$, $k=48$ is the hidden dimension, and $\textbf{b}\in \mathbb{R}^{2}$ is the bias vector.
For student $u$ answering item $i_{u,t}$ at time $t$, the prediction loss $l_{u,t}$ can be modeled with the binary cross entropy loss:
\begin{equation} 
    l_{u,t} = \textbf{r}_{u,v_c,t} \log(\textbf{p}_{u,t} + (\textbf{1} - \textbf{r}_{u,v_c,t}) \log( \textbf{1} - \textbf{p}_{u,t}))\color{black},\\
\end{equation}
where \textbf{1} is an all-one vector and $\textbf{r}_{u,v_c,t}$ is the one-hot encoding vector of the binary response ${r}_{u,v_c,t}$. 
Finally, the total loss $\mathcal{L}_{KT}$ for knowledge tracing (KT) can be represented as
\begin{equation} 
    \mathcal{L}_{KT} = \sum_{u \in \Omega} \sum_{t}^{L_u} l_{u,t},
\end{equation}
where $L_u$ is the length of the time series sequence $X_u$.
% for student $u$.

\subsubsection{Outcome Prediction}
\label{sec:outcomeprediction}
In addition to tracing students' knowledge state, we consider both video interaction and forum-activity to form a complete knowledge representation for outcome prediction~\cite{https://doi.org/10.48550/arxiv.2208.01182}.
For each student $u$, the complete activity record at time step~$t$ is defined as $\textbf{a}_{u,t} = \mathbbm{1}(c) \oplus \mathbbm{1}(v_c) \oplus \mathbbm{1}(r_{u,v_c,t}) \oplus \mathbbm{1}(f_{u,t})$, where $f_{u,t}$ represents the forum-participation activity $f_u$ made by student $u$ at time $t$.
% We concatenate video interaction and forum-participation activity in this way to ensure a consistent activity dimension. 
We adopt this concatenation of video interaction and forum-participation activity to maintain uniformity in the activity dimension.
% Note that video and forum activities do not happen simultaneously; whenever a student watches a video, the forum-participation part of this encoding, i.e., $\mathbbm{1}(f_{u,t}) = \mathbf{0}$. 
Note that video and forum activities occur separately; whenever a student watches a video, the forum-participation part of this encoding, i.e., $\mathbbm{1}(f_{u,t})$, is set as $\mathbf{0}$.
Likewise, whenever a student makes a forum-participation activity, the video part of $\textbf{a}_{u,t}$, i.e., $[\mathbbm{1}(v_c) \oplus \mathbbm{1}(r_{u,v_c,t})] = \mathbf{0}$.

{
Given a series of activities $A_u = \{ \textbf{a}_{u,1}, \textbf{a}_{u,2}, ..., \textbf{a}_{u,t_u} \}$, the goal of outcome prediction is to infer a binary classification label $s_u \in \{0, 1\}$ indicating whether the student successfully completed the course or not. Note that $t_u$, the total number of activities for student $u$ used for modeling, can vary based on the time-frame of interest for the prediction tasks; it can represent any segment of the course duration, such as the entire course, or segments more appropriate for early detection prediction, e.g., the first couple weeks of the course. This flexibility allows for tailored analyses at different stages of the course, accommodating various intervals of student engagement. The specific designations of $t_u$ used in our experiments will be detailed in the experiments section.}
\color{black}
We follow \cite{https://doi.org/10.48550/arxiv.2208.01182}'s suggestion and leverage attention-based Gated Recurrent Units (GRU) to capture dependencies over long time periods~\cite{Chu2021ClickBasedSP}.
The model takes encoded activities as an input, generates learned representations for each student, and predicts learning outcome. 
The hidden state of the GRU model is defined as follows:
\begin{equation}
\textbf{h}_{u,t} = \textup{GRU}\left(\textbf{a}_{u,t}, \; \textbf{h}_{u,t-1}\right).
\end{equation}
% Encoding a long time sequence $\textbf{a}_{u,t}$ into a single final state $\textbf{h}_{u,t}$ might suffer from significant information loss.
% To overcome this,~\cite{Vaswani2017AttentionIA} proposed a self-attention mechanism that weights the $\textbf{h}_{u,t}$ over time.
Encoding a long temporal sequence $\textbf{a}_{u,t}$ into a single final state $\textbf{h}_{u,t}$ can lead to substantial loss of information. To address this concern, ~\cite{Vaswani2017AttentionIA}  introduced a self-attention mechanism that assigns weights to $\textbf{h}_{u,t}$ across time. 
% Specifically, instead of forcing the network to encode all the information into the final state, an attention module takes all the $\textbf{h}_{u,t}$ as inputs and generates the learned representation $\widetilde{\textbf{h}}_u$ as an output.
Rather than forcing the network to condense all information into the final state, an attention module accepts all $\textbf{h}_{u,t}$ as input and produces the learned representation $\widetilde{\textbf{h}}_u$ as output.
Applying this concept to our context, we define an attention module as:
\begin{equation} 
    \widetilde{\textbf{h}}_u = \sum_{t} \alpha_t \textbf{h}_{u,t},\\
\end{equation}
where the weight $\alpha_t = \frac{\exp \left(e_t\right)}{\sum_t \exp\left(e_t\right)}$, $e_t = p_t^\top \tanh\left(\textbf{W}_{ \alpha}\textbf{h}_{u,t}\right)$, and $\textbf{W}_{\alpha}$ is a learned parameter.
% A linear layer then transforms the learned representation, and the predicted probability of pass/fail is obtained as:
Then, a linear layer converts the representation into the predicted pass/fail probability:
\begin{equation} 
    \textbf{s}'_u = \textup{softmax}\left(\textbf{W} \cdot \widetilde{\textbf{h}}_u + \textbf{b}\right),\\
\end{equation}
where $\textbf{W}\in \mathbb{R}^{k \times 2}$ is the weight matrix for linearly transforming $\widetilde{\textbf{h}}_u$, $k=48$ is hidden dimension, and $\textbf{b} \in \mathbb{R}^{2}$ is the bias vector.
The loss $\mathcal{L}_{OP}$ for assessing the quality of outcome prediction (OP) is defined as binary cross-entropy loss:
\begin{equation} 
    \mathcal{L}_{OP} = -\sum_{u \in \Omega} \textbf{s}_u \log(\textbf{s}'_u) + (\textbf{1} - \textbf{s}_u) \log( \textbf{1} - \textbf{s}'_u),\\
\end{equation}
% where \textbf{1} is an all-one vector, $\textbf{s}'_u$ is the prediction of the model, and $\textbf{s}_u$ is the one-hot encoding vector of the binary label $s_u$. 
where $\textbf{s}'_u$ is the model's prediction, \textbf{1} is an all-one vector, and $\textbf{s}_u$ is the one-hot encoded vector of the binary label $s_u$.
\vspace{-0.1in}
\section{Experimental Evaluation}
\label{sec:exp}
We now carry out experiments on our three online course datasets from Section~\ref{sec:tasks} to evaluate our personalization methodology MLPFL. 
We employ the standard AUC metric for evaluation. See Appendix~\ref{sec:appendix-implementation} for implementation details.

\subsection{Models and Baselines}
\label{sec:mdlbase}
We compare MLPFL against baselines in several different configurations.
These configurations differ in terms of several attributes: (i) scenario (i.e., scenario I (\texttt{sc1}) and II  (\texttt{sc2}) we introduced in Section~\ref{sec:sc1_course} and~\ref{sec:sc2_demo}), (ii) architecture (i.e., local (\texttt{L}), global (\texttt{G}), and personalized (\texttt{P}) model), (iii) FL aggregation method (i.e., average-based aggregation (\texttt{AV}), attention-based aggregation (\texttt{AT})), and (iv) hierarchy information (for the scenario with demographic personalization).
For conciseness, we use a general structure: 
% \vspace{-0.2in}
\begin{equation} 
\begin{split}
\texttt{Method} =  &[ \texttt{<Scenario>-<Architecture>-}\\
&\texttt{<Aggregation>-<Hierarchy>}]
\end{split}
\end{equation}
% \vspace{-0.2in}
% \begin{equation} 
% \begin{aligned}
%     \texttt{Method} =  &[ \texttt{<Scenario>-<Architecture>-\\
%             <Aggregation>-<Hierarchy>}]
% \end{aligned}
% \end{equation}
to represent each baseline  according to the attributes listed in Table~\ref{tb:config}.
% \texttt{<Hierarchy>} appears for scenario II to indicate which level of the hierarchy -- bottom (\texttt{B}), middle (\texttt{M}), top (\texttt{T}) -- is used for the final adapted model.
% Bottom, middle, and top represent models for demographic subgroups, course-specific models, and the global model, respectively.
In Scenario II, \texttt{<Hierarchy>} specifies the level within our hierarchical adaptation structure where the model is taken from--bottom (B), middle (M), or top (T). As discussed in Sec. II-B, “B” models are tailored to demographic subgroups within courses, providing the most fine-granular adaptation. “M” describes course-specific models that are adapted based on the learning patterns of students within a particular course. “T” indicates the global model, which is aggregated across all courses and demographics to offer the most coarse-granular yet comprehensive overview. This hierarchical approach enables nuanced modeling that ranges from highly personalized to broadly generalized analyses. \color{black} For example, \texttt{[sc2-P-AT-B]} corresponds to the personalized model that adapts the attention-aggregated global model based on the data of the bottom level (demographic subgroups) in scenario II.

Also, under the \texttt{<Architecture>} attribute, note that both the “G” and “P” architectures have global federated models, despite the “global (G)” name. The difference is that the global federated models in the “G” category are obtained through aggregations in standard, non-personalized federated learning, whereas those in the “P” category are defined according to our meta-learning adaptation procedures formalized in Section~\ref{sec:method}. Thus, the “G” category contains important baselines for evaluating our meta-learning-based student modeling. We will clarify the exact procedure followed in each algorithm setup from Table~\ref{tb:config}.
\color{black}

\begin{table}
\caption{Illustration of algorithm setups: \texttt{sc1} and \texttt{sc2} denote the first and second scenarios, and \texttt{L}, \texttt{G}, and \texttt{P} stand for local, global, and personalized models. \texttt{<Aggregation>} includes \texttt{AV} for average-based and \texttt{AT} for attention-based aggregations. In scenario II, three hierarchy levels—demographic (\texttt{B}), course (\texttt{M}), and global (\texttt{T})—are used.}

\begin{center}
\scalebox{1}{
\begin{tabular}{lc}
\hline
{Configuration} & {Attribute} \\\hline
\texttt{<Scenario>} & \texttt{sc1}, \texttt{sc2}\\
\texttt{<Architecture>} & \texttt{L}, \texttt{G}, \texttt{P} \\
\texttt{<Aggregation>} & \texttt{AV}, \texttt{AT} \\
\texttt{<Hierarchy>} (optional) & \texttt{B}, \texttt{M}, \texttt{T} \\

\hline
\end{tabular}}
\end{center}
% \vspace{-3mm}
\vspace{-0.3in}

\label{tb:config}
\end{table}
% local (L), global (G), and personalized (P) baselines for each scenario.

\subsubsection{Algorithms for scenario I (\texttt{<Scenario> = sc1})}
\textbf{{Local Modeling (\texttt{<Architecture> = L})}} We construct a local baseline \texttt{[sc1-L]} by building several local models based on student data in each course. More specifically, we train three local models, one for each course $c \in \mathcal{C}$, using datasets $\Omega^{\mathsf{train}}_c$ and evaluate each model on $\Omega^{\mathsf{test}}_c$.

\textbf{{Global Modeling (\texttt{<Architecture> = G})}} \color{black}
We implement three global models for scenario I. One of them, \texttt{[sc1-G]}, is a centralized model without FL, while the others, \texttt{[sc1-G-AV]} \& \texttt{[sc1-G-AT]}, are Global FL models.

$\bullet$ \texttt{[sc1-G]}: A centralized global model is trained on all the students' training data $\Omega^{\mathsf{train}}$ collected from all the courses and evaluated on $\Omega^{\mathsf{test}}$.

$\bullet$ \texttt{[sc1-G-AV]}: A federated global model is implemented from FedAvg~\cite{McMahan2017CommunicationEfficientLO}. We train several local models $\Theta_c^{(k,e)}$ on $\Omega^{\mathsf{train}}_c$, without meta-learning.
% After $E$ iterations, FedAvg weighs each local model based on the number of students in each course to conduct global aggregation:
Upon completing $E$ iterations, FedAvg aggregates local models, considering the student count in each course as the weighting factor for the aggregation:
\begin{equation} 
\Theta_{\mathsf{g}}^{(k+1)} = \sum_{c\in\mathcal{C}} \frac{N_c}{N} \Theta^{(k,E)}_c,
\end{equation} 
where $k$ is training round, $N_c$ is the number of students of course $c$, and $N=\sum_{c\in\mathcal{C}}N_c$.
After $K$ global aggregations, we take $\Theta_{\mathsf{g}}^{(K)}$ as \texttt{[sc1-G-AV]} to evaluate FedAvg on $\Omega_{c}^{\mathsf{test}}$.

$\bullet$ \texttt{[sc1-G-AT]}: An attention-based federated global algorithm is implemented from  FedAttn~\cite{Ji2019LearningPN}. After $E$ local iterations, FedAttn aggregates local models based on the attention mechanism introduced in (\ref{eq:globalatt}).
After $K$ rounds, the global model $\Theta_{\mathsf{g}}^{(K)}$ defined in (\ref{eq:globalupdate}) is evaluated on $\Omega_{c}^{\mathsf{test}}$.

% The centralized global model \texttt{[sc1-G]} is trained on all the students' training data $\Omega^{\mathsf{train}}$ collected from all the courses and evaluated on $\Omega^{\mathsf{test}}$.
% \texttt{[sc1-G-AV]} is implemented from FedAvg~\cite{McMahan2017CommunicationEfficientLO}.
% We first train several local models $\Theta_c^{(k,e)}$ on $\Omega^{\mathsf{train}}_c$ without meta-learning for training round $k$.
% After $E$ iterations, FedAvg weighs each local model based on the number of students in each course to conduct a standard global aggregation:
% \begin{equation} 
% \Theta_{\mathsf{g}}^{(k+1)} = \sum_{c\in\mathcal{C}} \frac{N_c}{N} \Theta^{(k,E)}_x,
% \end{equation} 
% where $N_c$ is the number of students within course $c$, and $N=\sum_{c\in\mathcal{C}}N_c$.
% After $K$ global aggregations, we then use $\Theta_{\mathsf{g}}^{(K)}$ as \texttt{[sc1-G-AV]} to evaluate FedAvg on $\Omega_{c}^{\mathsf{test}}$. 

% For \texttt{[sc1-G-AT]}, we implement an attention-based federated learning algorithm FedAttn~\cite{Ji2019LearningPN}.
% After $E$ local iterations, FedAttn aggregates local models based on the attention mechanism with the weights introduced in (\ref{eq:globalatt}).
% After $K$ global aggregations, the aggregated global model $\Theta_{\mathsf{g}}^{(K)}$ defined in (\ref{eq:globalupdate}) is evaluated on $\Omega_{c}^{\mathsf{test}}$.

\textbf{Adaptive \color{black} Modeling (\texttt{<Architecture> = P})} We consider two versions of our local model adaptation in FL for scenario I: \texttt{[sc1-P-AV]} and \texttt{[sc1-P-AT]}.
Both of these models adapt the global model based on meta-updates from the data in each course.

$\bullet$ \texttt{[sc1-P-AV]}: This follows the algorithm we introduced in Section~\ref{sec:sc1_course}, while replacing our attention-based aggregation method in (\ref{eq:globalatt}) and (\ref{eq:globalupdate}) with the averaging-based aggregation of PerFed~\cite{Fallah2020PersonalizedFL}.

$\bullet$ \texttt{[sc1-P-AT]}: This is the full version of our course-based adaptive method from Section~\ref{sec:sc1_course}.

% \begin{itemize}
% \item \texttt{[sc1-P-AV]}: It follows the algorithm we introduced in Section~\ref{sec:sc1_course}, while replacing our attention-based aggregation method in (\ref{eq:globalatt}) and (\ref{eq:globalupdate}) with the averaging-based aggregation of PerFed~\cite{Fallah2020PersonalizedFL}. 
% \item \texttt{[sc1-P-AT]}: The full version of our course-based personalized method introduced in Section~\ref{sec:sc1_data}-~\ref{sec:scen1_agg}.
% \end{itemize}

% \texttt{[sc1-P-AV]} replaces our attention-based aggregation method in (\ref{eq:globalatt}) and (\ref{eq:globalupdate}) with the averaging-based aggregation of PerFed~\cite{Fallah2020PersonalizedFL}. 
% We denote the training procedure we introduce in Section~\ref{sec:person_course} and \ref{sec:scen1_agg} as \texttt{[sc1-P-AV]}.

\subsubsection{Algorithms for scenario II (\texttt{<Scenario> = sc2})}
\textbf{Local Modeling \texttt{(<Architecture> = L)}}
In scenario II, the bottom hierarchy is grouped by demographic variables. Therefore, we build a local baseline \texttt{[sc2-L]} with a separate model for each of these demographic subgroups.
Each model in \texttt{[sc2-L]} is trained on student data $\Omega^{\mathsf{train}}_{\{c,I,x\}}$ for a specific subgroup $\{c, I, x\}$ (i.e., subgroup $x$ of demographic variable $I$ in course $c$) and evaluated on $\Omega^{\mathsf{test}}_{\{c,I,x\}}$.

\textbf{Global Modeling \texttt{(<Architecture> = G)}} 
We also consider a centralized global \texttt{[sc2-G]} and several federated global models \texttt{[sc2-G-AV-M]} \& \texttt{[sc2-G-AT-M]} \& \texttt{[sc2-G-AV-T]} \& \texttt{[sc2-G-AT-T]}.
For the federated global models, we first train several local models based on the demographic subgroups within each course. Then, we aggregate them to form different global models at each level (i.e., course(middle)-level global models (\texttt{[sc2-G-AV-M]} \& \texttt{[sc2-G-AT-M]}) and top level global models (\texttt{[sc2-G-AV-T]} \& \texttt{[sc2-G-AT-T]})).

$\bullet$ \texttt{[sc2-G]}: Architecturally, this is the same as \texttt{[sc1-G]} because they both use all students $\Omega^{\mathsf{train}}$ to build one model. However, we evaluate \texttt{[sc2-G]} on each demographic subgroup $\Omega^{\mathsf{test}}_{\{c,I,x\}}$ instead of each course.

$\bullet$ \texttt{[sc2-G-AV-M]}: In this case, the demographic models are first locally trained. Then, this baseline obtains the aggregated course(middle)-level model $\Theta_c^K$ at round $K$ using the weighted average method of FedAvg~\cite{McMahan2017CommunicationEfficientLO}:
\begin{equation} 
\label{eq:sc2-g-av-m}
\Theta_{\mathsf{c}}^{K+1} = \sum_{x\in\mathcal{X}} \frac{N_x}{N_c} \Theta^{(K,E)}_x,
\end{equation} 
where $\Theta^{(K,E)}_x$ is a demographic model trained locally, $N_x$ is the number of students within subgroup $x$, $N_c$ is the number of students within course $c$, and $N_c=\sum_{x\in\mathcal{X}}N_x$.

$\bullet$ \texttt{[sc2-G-AT-M]}: This is the course(middle)-level model $\Theta_{g}^{K+1}$ from (\ref{eq:course_agg}), that aggregates local demographic models via the attention-based aggregation method. 
% This is the model described in (\ref{eq:course_agg}) in Section~\ref{sec:scen2_agg}.

$\bullet$ \texttt{[sc2-G-AV-T]}: After obtaining the course(middle)-level global models from~(\ref{eq:sc2-g-av-m}), we aggregate them again to the top level global model. Specifically, \texttt{[sc2-G-AV-T]} is obtained by aggregating $\Theta_c^K$ as:
\begin{equation} 
\Theta_{g}^{K+1} = \sum_{c\in\mathcal{C}} \frac{N_c}{N} \Theta^{K+1}_c.
\end{equation} 

$\bullet$ \texttt{[sc2-G-AT-T]}: This applies the attention aggregation mechanism to \texttt{[sc2-G-AT-M]} and obtains the top-level global model based on (\ref{eq:global_agg}).

\textbf{Personalized Modeling \texttt{(<Architecture> = P)}} 
We consider five personalized algorithms in this scenario. One of the baselines, \texttt{FedIRT}, uses educational theory for the aggregation, while four --\texttt{[sc2-P-AV-M]}, \texttt{[sc2-P-AT-M]}, \texttt{[sc2-P-AT-B]}, and \texttt{[sc2-P-AT-B]}-- are different configurations of our MLPFL algorithm.
\texttt{[sc2-P-AV-M]} and \texttt{[sc2-P-AT-M]} are the course(middle)-level personalizations adapted from the top global model, while \texttt{[sc2-P-AV-B]} and \texttt{[sc2-P-AT-B]} are the corresponding demographic(bottom)-level personalizations adapted from the course(middle)-level models.

$\bullet$ \texttt{FedIRT}: This is based on the federated deep knowledge tracing (FDKT)~\cite{Wu2021FederatedDK} method discussed in Section~\ref{sec:introduction}. 
FDKT employs classical test theory and item response theory (IRT)~\cite{Tatsuoka1968StatisticalTO} to compute a score that reflects the ``data quality" of local subgroups, which then influences the weighting of each subgroup during model aggregation.
For our setting, we execute the model update and aggregation process by computing the IRT confidence $\alpha^{\mathsf{IRT}}_x$ for each student subgroup $x$ in scenario II, using the students' responses to the end-of-video quizzes.
Within each course, local models are trained for each demographic variable.
Following ~\cite{Wu2021FederatedDK}, local model updates use interpolation, where the initial model for training round $k$ is defined as: 
    \begin{equation} 
{\small \Theta^{(k,0)}_x = \lambda^{(k)}_x \Theta^{(k-1,E)}_x + (1-\lambda^{(k)}_x)\Theta^{(k)}_g, }
\end{equation} 
where 
\begin{equation} 
\small \lambda^{(k)}_x = \left({\Theta^{(k-1,E)}_x \cdot \Theta^{(k)}_{\mathsf{g}}}\right)\big/\left({\big\Vert\Theta^{(k-1,E)}_x\big\Vert \times\big\Vert\Theta^{(k)}_{\mathsf{g}}\big\Vert}\right), 
\end{equation} 
and the final local model $\Theta^{(k,E)}_x$ is obtained after $E$ local epochs of conventional gradient descent.
The aggregated global model is then computed as 
\begin{equation} 
{\small \Theta^{(k+1)}_{\mathsf{g}} = \sum_{x \in \mathcal{X}} \alpha^{\mathsf{IRT}}_x \Theta^{(k,E)}_x}.
\end{equation}

$\bullet$  \texttt{[sc2-P-AV-M]}: This is built by taking one meta-update from \texttt{[sc2-G-AV-T]}, i.e., following~(\ref{eq:metaUpd_demo}). This creates one personalized model for each course, using stratified sampling of students in each subgroup for each course.

$\bullet$ \texttt{[sc2-P-AT-M]}: This baseline is adapted from \texttt{[sc2-G-AT-T]} by taking one meta-update.
% , as in \texttt{[sc2-P-AV-M]}. 

$\bullet$ \texttt{[sc2-P-AV-B]}: This is constructed through one further meta-update step from \texttt{[sc2-P-AV-M]}.
% the course-level models in .
It uses the datasets from each demographic subgroup to create a separate model for each subgroup in each course.

$\bullet$ \texttt{[sc2-P-AT-B]}: Taking one further meta-update step from \texttt{[sc2-P-AT-M]}, \texttt{[sc2-P-AT-B]} corresponds to the full MLPFL method we introduced in Section \ref{sec:scen2_meta}- \ref{sec:scen2_agg}. 

\vspace{-0.1in}
\subsection{Experimental Results}
\label{sec:result}
In Tables~\ref{tb:sc1_kt} and \ref{tb:sc1_pp}, we compare the predictive quality of knowledge tracing and outcome prediction for each method across courses in scenario I.
For scenario II, Tables ~\ref{tb:sc2_kt_g1}, ~\ref{tb:sc2_kt_c1}, and ~\ref{tb:sc2_kt_a1} compare the performance of knowledge tracing on each student group, while Tables~\ref{tb:sc2_pp_g1}, ~\ref{tb:sc2_pp_c1}, and ~\ref{tb:sc2_pp_a1} in Appendix~\ref{sec:appendix-outcome} are the results for outcome prediction.
Additionally, Tables~\ref{tb:sc1_early2} and \ref{tb:sc1_early1} provide insights into early prediction performance using different time-frames of students' activities.

\color{black}
In this section, all results are shown  on the set of students who provided specific demographic information $I$; our models are trained on  $\Omega_{c,I}^{\mathsf{train}}$ and evaluated on each subgroup of $\Omega_{c,I}^{\mathsf{test}}$.
% Precisely, we divide the dataset into groups by gender, continent, and age information for the experiments in Tables~\ref{tb:sc2_kt_g1}\&\ref{tb:sc2_pp_g1}, \ref{tb:sc2_kt_c1}\&\ref{tb:sc2_pp_c1}, and~\ref{tb:sc2_kt_a1}\&\ref{tb:sc2_pp_a1}, respectively.
More specifically, for the experiments in Tables~\ref{tb:sc2_kt_g1}\&\ref{tb:sc2_pp_g1}, \ref{tb:sc2_kt_c1}\&\ref{tb:sc2_pp_c1}, and~\ref{tb:sc2_kt_a1}\&\ref{tb:sc2_pp_a1}, the dataset is partitioned into groups according to gender, continent, and age, respectively.
% We defer the results on incorporating students who did not disclose their personal information (i.e., the models trained on $\{\Omega_{c,I}^{\mathsf{train}}, \Omega_{\neg I}^{\mathsf{train}}\}$ and tested on each subgroup of $\{\Omega_{c,I}^{\mathsf{test}}, \Omega_{\neg I}^{\mathsf{test}}\}$) to Appendix~\ref{sec:appendix}; the results are qualitatively similar.
We present the results for students who did not disclose their personal information (i.e., the models trained on $\{\Omega_{c,I}^{\mathsf{train}}, \Omega_{\neg I}^{\mathsf{train}}\}$ and tested on each subgroup of $\{\Omega_{c,I}^{\mathsf{test}}, \Omega_{\neg I}^{\mathsf{test}}\}$) to Appendix~\ref{sec:appendix}; these results exhibit similar qualitative patterns. 

\subsubsection{Discussion for scenario I}
\label{sec:disc_I}
From Tables~\ref{tb:sc1_kt} and \ref{tb:sc1_pp}, we see that in many cases, there is not a large difference in performance between the locally trained \texttt{[sc1-L]} for each course and the centrally trained \texttt{[sc1-G]} across courses. This is a key motivation for course adaptation: despite \texttt{[sc1-G]} containing significantly larger training data, it is not preserving course-specific information. 
On the other hand, \texttt{[sc1-G]} generally outperforms both versions of global FL (\texttt{[sc1-G-AV]} and \texttt{[sc1-G-AT]}), indicating that centralized training across the entire dataset is more effective than decentralized models in scenarios without adaptation.

Across Tables~\ref{tb:sc1_kt} and~\ref{tb:sc1_pp}, our proposed adaptive models (\texttt{[sc1-P-AV]} and \texttt{[sc1-P-AT]}) obtain anywhere from 15-25\% improvement in AUC over the non-adaptive models, depending on the course, aggregation, and downstream task.
The improvements obtained by the adaptive models show the significance of adapting the global models based on local students within each course.
Compared to \texttt{[sc1-P-AV]}, MLPFL (\texttt{[sc1-P-AT]}) performs better on most courses, confirming our intuition on the advantages of layer-wise model aggregations with an attention mechanism.
\textit{Importantly, these findings suggest that even without demographic information, MLPFL is able to obtain substantial improvement in student modeling through course-level personalizations.}

\begin{table}[tp]
\caption{AUC results for knowledge tracing of different models}

\begin{center}

\scalebox{0.9}{

\begin{tabular}{cccc}
\hline
\multicolumn{1}{c}{Model} & FOC & \multicolumn{1}{c}{QDS} & MOSFETs\\ \hline\hline
\texttt{[sc1-L]}  & .533 (.013) & .537 (.009) & .529 (.011) \\\hline
\texttt{[sc1-G]} & .528 (.007)& .529 (.010) & .531 (.011)\\
\texttt{[sc1-G-AV]} & .521 (.010)& .517 (.009) & .529 (.011) \\
\texttt{[sc1-G-AT]} & .526 (.008) & .525 (.012) & .519 (.007)\\
\hline
 \texttt{[sc1-P-AV]} & .614 (.016) & .618  (.009) & .603 (.014)\\
\texttt{[sc1-P-AT]}(MLPFL) & \textbf{.621 (.019)}& \textbf{.625 (.015)} & \textbf{.619 (.014)}\\
\hline
\end{tabular}
}
\end{center}
\vspace{-0.2in}

\label{tb:sc1_kt}
\end{table}

\begin{table}[tp]
\caption{AUC results for outcome prediction of different models.}

\begin{center}

\scalebox{0.9}{

\begin{tabular}{cccc}
\hline
\multicolumn{1}{c}{Model} & FOC & \multicolumn{1}{c}{QDS} & MOSFETs\\ \hline\hline
\texttt{[sc1-L]} & .553 (.021) & .558 (.017) & .549 (.019) \\\hline
\texttt{[sc1-G]} & .547 (.017)& .559 (.010) & .553 (.013)\\
\texttt{[sc1-G-AV]} & .550 (.013)& .546 (.017) & .551 (.009) \\
\texttt{[sc1-G-AT]} & .548 (.018) & .556 (.011) & .548 (.015)\\
\hline
\texttt{[sc1-P-AV]} & .678 (.020) & .689  (.011) & .658 (.017)\\
\texttt{[sc1-P-AT]}(MLPFL) & \textbf{.684 (.014)}& \textbf{.701 (.021)} & \textbf{.679 (.015)}\\
\hline
\end{tabular}
}
\end{center}
\vspace{-0.1in}
% \vspace{-0.35in}

\label{tb:sc1_pp}
\end{table}

\begin{figure}

\centering
\setlength{\abovecaptionskip}{1mm}
\vspace{-0.1in}
\includegraphics[width=0.7\linewidth]{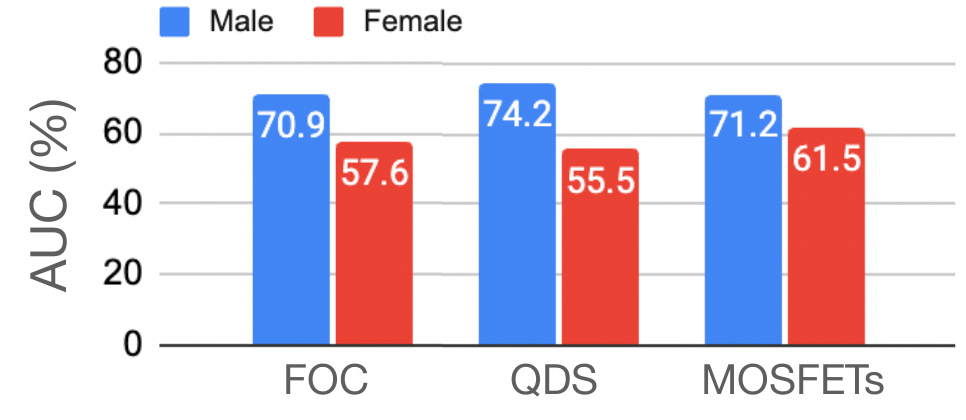}
\caption{ 
{Breakdown of outcome prediction performance by gender for Scenario I using our method, MLPFL (\texttt{[sc1-P-AT]}).}\color{black}} 
\vspace{-0.2in}
\label{fig:sc1_bias}
\end{figure}

\begin{table}[tp]
\caption{AUC results for early prediction using the first 3 weeks of data, across different models.\color{black}}

\begin{center}

\scalebox{1.0}{

\begin{tabular}{cccc}
\hline
\multicolumn{1}{c}{Model} & FOC & \multicolumn{1}{c}{QDS} & MOSFETs\\ \hline\hline
\texttt{[sc1-L]} & .462 & .459 & .488 \\\hline
\texttt{[sc1-G]} & .473 & .487 & .499 \\
\texttt{[sc1-G-AV]} & .470 & .469 & .489 \\
\texttt{[sc1-G-AT]} & .479  & .478 & .500\\
\hline
\texttt{[sc1-P-AV]} & .498 & .503   & .518\\
\texttt{[sc1-P-AT]}(MLPFL) & \textbf{.519}& \textbf{.534} & \textbf{.537}\\
\hline
\end{tabular}
}
\end{center}
\vspace{-0.2in}
% \vspace{-0.35in}

\label{tb:sc1_early2}
\end{table}
\vspace{-0.1in}

\begin{table}[tp]
\caption{AUC results for early prediction using half of each student’s data, across different models.\color{black}}

\begin{center}

\scalebox{1.0}{

\begin{tabular}{cccc}
\hline
\multicolumn{1}{c}{Model} & FOC & \multicolumn{1}{c}{QDS} & MOSFETs\\ \hline\hline
\texttt{[sc1-L]} & .487 & .473 & .491 \\\hline
\texttt{[sc1-G]} & .499 & .508 & .489 \\
\texttt{[sc1-G-AV]} & .503 & .486 & .512 \\
\texttt{[sc1-G-AT]} & .498  & .511 & .505\\
\hline
\texttt{[sc1-P-AV]} & .539 & .553   & .549\\
\texttt{[sc1-P-AT]}(MLPFL) & \textbf{.568}& \textbf{.592} & \textbf{.583}\\
\hline
\end{tabular}
}
\end{center}
% \vspace{-0.1in}
\vspace{-0.35in}

\label{tb:sc1_early1}
\end{table}

 Tables~\ref{tb:sc1_kt} and \ref{tb:sc1_pp} demonstrate that our method outperforms other baselines in two downstream tasks. We further analyze the breakdown of performance by demographic, particularly when demographic information is not considered in the modeling. Figure~\ref{fig:sc1_bias} shows the outcome prediction performance by gender across three courses using our method. Although our method generally improves performance for each course, significant variations are observed between demographic subgroups. This observation motivates the need for Scenario II, which incorporates demographic information into the modeling to address these disparities more effectively.
 \color{black}

% {To evaluate if our method can mitigate biases for minorities, particularly in the MOSFETs course—which has the smallest amount of student data in Scenario I—Table~\ref{tb:sc1_improve} summarizes the prediction improvements of each method over the population-level method (global modeling) for the MOSFETs course. The improvement of our method for the MOSFETs course across two student prediction modeling tasks demonstrates that our method can benefit minorities through personalization.
% }
% \color{black}

% We also expanded our analysis to include early prediction tasks. Specifically, we focused on the initial activities of each student by utilizing data from the first half of their course engagement (from $t = 1$ to $t = 0.5T_u$), aiming to predict their learning outcomes based on this early data. Table \ref{tb:sc1_early} presents the results of this extension, illustrating a slight decrease in AUC scores for each method when compared to using the full duration of data. However, our MLPFL continues to outperform other baselines in this early prediction context, demonstrating its effectiveness even with limited data.
In addition to the outcome prediction that utilizes all activities from each student throughout the entire course (Table~\ref{tb:sc1_pp}), we have expanded our analysis to include early prediction tasks. Specifically, we focused on two time-frames: (a) using only the first half of each student's recorded interactions in the course (Table~\ref{tb:sc1_early1}) and (b) using only the first three weeks' worth of data, out of approximately 17 weeks total for each course (Table~\ref{tb:sc1_early2}). Compared with using the full duration of data, the results show a slight decrease in AUC scores for each method. However, our MLPFL continues to outperform other baselines in these early prediction scenarios, demonstrating its effectiveness even with limited data.

\color{black}

\subsubsection{Discussion for scenario II}
\label{sec:dis_II}

\begin{figure*}[!ht]
    \centering
    \setlength{\abovecaptionskip}{1mm}
    \includegraphics[width=0.88\linewidth]{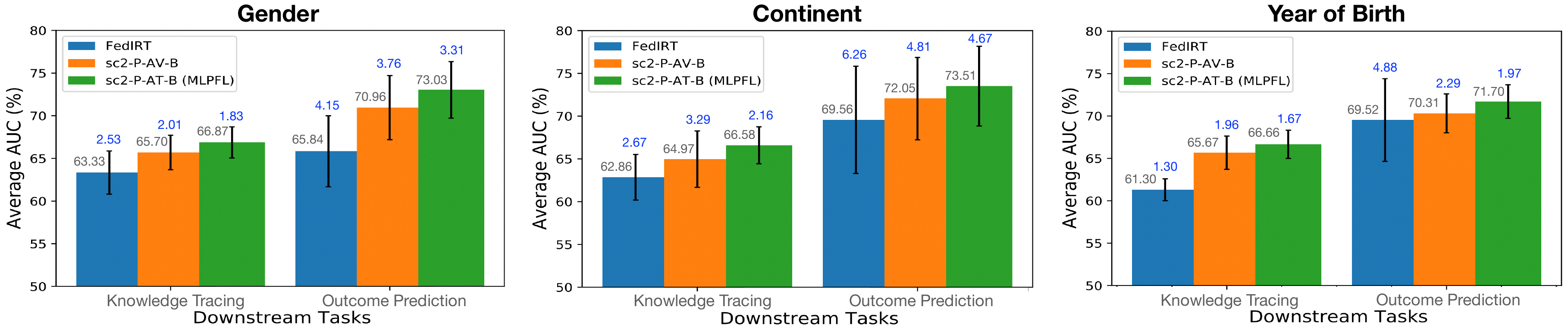}
	 \caption{ 
  {Mean and standard deviation (in blue text) AUC scores of personalized methods (\texttt{FedIRT}, \texttt{[sc2-P-AV-B]}, \texttt{[sc2-P-AT-B]}) across all subgroups in three courses. The increased mean AUC scores and the reduced variances show that MLPFL improves prediction quality while mitigating biases.}\color{black}} 
\vspace{-0.2in}

\label{fig:mean_var_info}
\end{figure*}

In Tables~\ref{tb:sc2_kt_g1}, \ref{tb:sc2_kt_c1}, and \ref{tb:sc2_kt_a1}, for knowledge tracing, we observe that the local model \texttt{[sc2-L]} provides on average 54\% AUC on each subgroup according to different demographic groupings, and in Tables~\ref{tb:sc2_pp_g1},~\ref{tb:sc2_pp_c1}, and ~\ref{tb:sc2_pp_a1}, we find on average 60\% AUC for outcome prediction. 
While providing moderate improvements over \texttt{[sc1-L]}, these results confirm our hypothesis that small subgroups do not have enough data to train high accuracy models individually. 
On the other hand, unlike in scenario I, the centralized global model \texttt{[sc2-G]} does not necessarily outperform the global federated models (\texttt{[sc2-G-AV-M]}\&\texttt{[sc2-G-AT-M]} and \texttt{[sc2-G-AV-T]}\&\texttt{[sc2-G-AT-T]}).
This is particularly true for underrepresented subgroups.
For example, in Table~\ref{tb:sc2_pp_g1}, for the MOSFETs dataset, \texttt{[sc2-G]} obtains the highest performance for the male subgroup among the non-personalized models, while two of the global FL models noticeably outperform it on the female subgroup.
As another example, in Table~\ref{tb:sc2_pp_c1}, for the QDS dataset, \texttt{[sc2-G-AT-M]} outperforms \texttt{[sc2-G]} noticeably for all subgroups except North America.
By incorporating such subgroup heterogeneity, the personalized models are able to obtain improvements across datasets and subgroups, as we discuss further next.

\begin{table*}[htp]
\caption{The performance on knowledge tracing obtained by different models on each gender subgroup.}

\begin{center}

\scalebox{0.9}{

\begin{tabular}{lccccccc}
\hline
\multicolumn{1}{c|}{Dataset}& \multicolumn{2}{c|}{FOC} & \multicolumn{2}{c|}{QDS}  & \multicolumn{2}{c}{MOSFETs}\\  \hline
\multicolumn{1}{c|}{Model} & male & \multicolumn{1}{c|}{female} & male & \multicolumn{1}{c|}{female} & male & female \\ \hline\hline
\texttt{[sc2-L]} & .532 (.009) & .519 (.004) & .541 (.010) & .523 (.006) & .538 (.011) & .519 (.003) \\\hline
\texttt{[sc2-G]} & .539 (.005) & .557 (.012) & .538 (.008) & .543 (.006) & .567 (.009) & .551 (.010) 
\\
\texttt{[sc2-G-AV-M]}  & .521 (.003) & .539 (.008) & .547 (.010) & .551 (.013) & .543 (.011) & .538 (.008) \\
\texttt{[sc2-G-AT-M]} & .529 (.002) & .534 (.012) & .552 (.013) & .546 (.015) & .557 (.017) & .568 (.020) \\
\texttt{[sc2-G-AV-T]} & .517 (.002) & .526 (.008) & .510 (.004) & .527 (.009) & .530 (.006) & .526 (.011) \\
\texttt{[sc2-G-AT-T]} & .523 (.008) & .515 (.006) & .523 (.013) & .538 (.018) & .549 (.023) & .530 (.017) \\
\hline
\texttt{FedIRT} & .667 (.018) & .652 (.025) & .621 (.023) & .667 (.022) & .609 (.015) & .598 (.011)  \\
\texttt{[sc2-P-AV-M]} & .665 (.029) & .630 (.024) & .649 (.031) & .651 (.028) & .624 (.022) & .602 (.018)  \\
\texttt{[sc2-P-AT-M]}(MLPFL) & .680 (.033) & .633 (.028) & .655 (.024) & .668 (.031) & .637 (.027) & .620 (.019) \\
\texttt{[sc2-P-AV-B]} & .679 (.033) & \textbf{.661 (.028)} & .658 (.026) & .671 (.031) & .630 (.017) & .625 (.019) \\
\texttt{[sc2-P-AT-B]}(MLPFL)& \textbf{.692 (.031)} & .654 (.028) & \textbf{.671 (.028)} & \textbf{.684 (.024)} & \textbf{.644 (.021)} & \textbf{.631 (.016)} \\
\hline

\end{tabular}
}
\end{center}
% \vspace{-0.1in}
% \vspace{-0.3in}
\vspace{-0.25in}

\label{tb:sc2_kt_g1}
\end{table*}

For both the knowledge tracing task (Tables~\ref{tb:sc2_kt_g1},~\ref{tb:sc2_kt_c1}, and~\ref{tb:sc2_kt_a1}) and the outcome prediction task (Tables~\ref{tb:sc2_pp_g1},~\ref{tb:sc2_pp_c1}, and~\ref{tb:sc2_pp_a1}), \textit{our MLPFL obtains substantial improvements over the non-personalized models on each demographic subgroup}. Specifically, for the FOC and QDS datasets, we obtain 15-35\% improvement in AUC over the highest performing non-personalized model across subgroups. 
The improvements obtained by the personalized models are noticeable but more limited in the MOSFETs dataset, possibly due to its overall smaller size (see Table~\ref{tb:course}).
For both tasks, MLPFL consistently outperforms \texttt{FedIRT} except in 4 out of approximately 50 subgroup cases across courses, underscoring the advantages of our meta learning-based personalization strategy.
% This improvement validates our hypothesis that weighting student subgroups based on their ``data quality'' as in \texttt{FedIRT} may lose important information on subgroup heterogeneity, which is captured by our redefinition in~(\ref{eq:demoLoss}).
This enhancement confirms our premise that assigning weights to student subgroups according to their ``data quality'' as done in \texttt{FedIRT} might result in the loss of crucial information regarding subgroup diversity, a concern addressed by our redefinition in~(\ref{eq:demoLoss}).
Moreover, \texttt{FedIRT} does not consider course-specific modeling, whereas our personalized architecture adapts the student models across different courses and demographics.

These results also show how our methodology benefits from the personalization hierarchy. Specifically, in conducting the meta-updates that move from \texttt{[sc2-G-AV-T]} to \texttt{[sc2-P-AV-M]} and \texttt{[sc2-G-AT-T]} to \texttt{[sc2-P-AT-M]}, we find that the course-level personalized models outperform the global models by at least 15\% AUC for both downstream tasks.
By considering the lowest-layer subgroupings and performing additional meta-updates that move from \texttt{[sc2-P-AV-M]} to \texttt{[sc2-P-AV-B]} and \texttt{[sc2-P-AT-M]} to \texttt{[sc2-P-AT-B]}, demographic-level personalized models further improve the AUC by 7\%-9\% on the FOC and QDS datasets over course-level personalized models.
Bulding upon our MLPFL results for scenario I, this shows how our personalization methodology benefits from increasing amounts of information available about students, whereas the non-personalized FL models do not show this trend when moving from top to middle layer modeling.

To directly assess the bias mitigation in Scenario II, in Table~\ref{tb:sc2_improve}, we show the standard deviation in AUC (expressed as a percentage of the mean) across each demographic group obtained by different algorithms. A model with less bias for a demographic variable will have lower standard deviation across the subgroup, as it indicates closer performance achieved across the different categories. Compared to the globally trained {\tt [sc1-G]} model, we see that our algorithms in Scenario I actually slightly increase the standard deviation across demographic groups, since these characteristics are not explicitly accounted for. In contrast, our algorithms in Scenario II lead to substantial reductions in standard deviation for all demographic groups in each of the courses and tasks compared to the global model (65-70\%). Thus, when furnished with demographic information, our approach can both mitigate prediction biases for minority groups while leading to improvements in performance for all groups.
\color{black}

% Finally, compared to \texttt{[sc2-P-AV-B]}, {\ourmethod} (\texttt{[sc2-P-AT-B]}) performs better on most student subgroups, confirming our intuition on the advantages of layer-wise model aggregations with an attention mechanism.

Finally, Figure~\ref{fig:mean_var_info} presents both the mean and variance in prediction quality for each demographic subgroup obtained by \texttt{FedIRT} and our bottom-layer MLPFL methodology, averaged across courses.
We can see that MLPFL consistently provides an improvement in mean AUC over \texttt{FedIRT}, with the attention mechanism providing an extra boost.
Importantly, we also see that \texttt{[sc2-P-AT-B]} reduces the AUC variance across subgroups in five of the six cases. The exception case is the one with the lowest variance across all methods.
\textit{The increase in mean and reduction in variance of prediction quality provided by MLPFL confirms that it lessens the impact of data subgroup availability biases in student modeling.}

\begin{table}[tp]
\caption{Prediction performance on knowledge tracing obtained with different models on each student subgroup grouped by continent. AS, AF, EU, NA, and SA represent Asian, African, European, North American, and South American subgroups.}

\begin{center}

\scalebox{0.72}{

\begin{tabular}{cccccc}
\hline
\multicolumn{1}{c|}{Dataset}& \multicolumn{5}{c}{FOC} \\  \hline
 \multicolumn{1}{c|}{Model} & AS & AF & EU & NA & SA \\ \hline\hline
\texttt{[sc2-L]} & .521 (.007) & .519 (.005) & .536 (.008) & .517 (.004) & .532 (.009)	
\\\hline

\texttt{[sc2-G]} & .537 (.009) & .546 (.013) & .535 (.007)  & .521 (.005) & .530 (.010)	
\\
\texttt{[sc2-G-AV-M]} & .555 (.009) & .562 (.013) & .559 (.011)  &	.538 (.007) & 	.556 (.013)
\\
\texttt{[sc2-G-AT-M]} & .578 (.015) & .563 (.016) & .557 (.013) &	.569 (.011) & .568 (.009)\\
\texttt{[sc2-G-AV-T]} & .546 (.007) & .532 (.011) & .543 (.013) &	.550 (.010) & .567 (.008)\\
\texttt{[sc2-G-AT-T]} & .557 (.016) & .549 (.018) & .561 (.020)  & .548 (.013)	& .557 (.015) \\ \hline
\texttt{FedIRT} & .601 (.007) & .637 (.019)& .605 (.010)& .649 (.016)& .633 (.015)\\
\texttt{[sc2-P-AV-M]} &.614 (.020) &.603 (.024) &.627 (.023)  &.600 (.018)	 & .611 (.016)\\
\texttt{[sc2-P-AT-M]}(MLPFL)  & .628 (.021) &  .617 (.026) &	.639 (.024) & .624 (.020) &.643 (.024)\\
\texttt{[sc2-P-AV-B]} & .640 (.023) & \textbf{.653 (.027)} &.635 (.013)  & \textbf{.648 (.017)} &.647 (.025)\\
\texttt{[sc2-P-AT-B]}(MLPFL) & \textbf{.652 (.023)} & .648 (.019) & \textbf{.667 (.015)} &	.643 (.023) & \textbf{.652 (.021)} \\

\hline\hline
\multicolumn{1}{c|}{Dataset}& \multicolumn{5}{c}{QDS} \\  \hline
 \multicolumn{1}{c|}{Model} & AS & AF & EU & NA & SA \\ \hline\hline
\texttt{[sc2-L]} & .510 (.004) & .533 (.006) & .548 (.008) &.564 (.010) &.523 (.007)
\\\hline

\texttt{[sc2-G]} & .556 (.006) & .533 (.011) & .548 (.012) &	.540 (.015) &	.536 (.010)
\\
\texttt{[sc2-G-AV-M]} & .547 (.016)& .583 (.009) &.546 (.013)  &.557 (.018)	 &	.589 (.021)
\\
\texttt{[sc2-G-AT-M]} & .583 (.013)& .554 (.011)& .562 (.015)  &	.567 (.019) & .565 (.016)\\
\texttt{[sc2-G-AV-T]} & .532 (.010)& .568 (.018)& .521 (.007) &.556 (.013) & .574 (.018)\\
\texttt{[sc2-G-AT-T]} & .564 (.015) & .576 (.014)& .543 (.016) &	.575 (.013) & .580 (.009)\\ \hline
\texttt{FedIRT} & .659 (.019) & .632 (.022)& .655 (.017)& .619 (.015)& .628 (.017)\\
\texttt{[sc2-P-AV-M]} & .627 (.025)& .604 (.019)& .632 (.016) &.628 (.020) &.617 (.023) \\
\texttt{[sc2-P-AT-M]}(MLPFL)  &.651 (.015) &.638 (.018) & .667 (.022) &	.662 (.024) & .654 (.016)\\
\texttt{[sc2-P-AV-B]} & .670 (.028)& .654 (.025)& .668 (.014) &.658 (.016)	 & \textbf{.677 (.017)}\\
\texttt{[sc2-P-AT-B]}(MLPFL) & \textbf{.698 (.018)} & \textbf{.671 (.015)} &\textbf{.683 (.020)}  &\textbf{.692 (.027)}	 &.674 (.023) \\

\hline\hline
\multicolumn{1}{c|}{Dataset}& \multicolumn{5}{c}{MOSFETs} \\  \hline
 \multicolumn{1}{c|}{Model} & AS & AF & EU & NA & SA \\ \hline\hline
\texttt{[sc2-L]} & .531 (.010) & .518 (.004)& .510 (.006) & .537 (.008)	 &.555 (.015)	
\\\hline

\texttt{[sc2-G]} & .528 (.013)& .537 (.018)&.550 (.015)  &.542 (.007)	 &.536 (.009)	
\\
\texttt{[sc2-G-AV-M]} &.548 (.015) &.532 (.007) &.569 (.006)  &.547 (.008)	 &	.563 (.015)
\\
\texttt{[sc2-G-AT-M]} &.569 (.017) &.583 (.016) &.552 (.010)  &.543 (.009)	 &.538 (.010)\\
\texttt{[sc2-G-AV-T]} & .532 (.010)& .518 (.015)& .547 (.009) &.535 (.004)	 & .533 (.013)\\
\texttt{[sc2-G-AT-T]} & .553 (.015)& .547 (.008)& .532 (.016) &	.514 (.005) & .528 (.008)\\ \hline
\texttt{FedIRT}  & .592 (.008) & .620 (.010)& \textbf{.617 (.015)}& .584 (.009)& .603 (.015)\\
\texttt{[sc2-P-AV-M]}  &.588 (.016) &.563 (.015) &.551 (.020)  &.549 (.018)	 & .543 (.012)\\
\texttt{[sc2-P-AT-M]}(MLPFL)   & .607 (.025)& .584 (.017)& .586 (.015) &	.553 (.010) & .567 (.015)\\
\texttt{[sc2-P-AV-B]}  & .624 (.018)& \textbf{.632 (.016)}&.615 (.020)  &.602 (.014)	 &.607 (.016) \\
\texttt{[sc2-P-AT-B]}(MLPFL)   &\textbf{.632 (.018)} &.618 (.012) & .600 (.023) &	\textbf{.613 (.015)} & \textbf{.629 (.020)}\\

\hline
\end{tabular}
}
\end{center}
\label{tb:sc2_kt_c1}
\vspace{-9mm}
\end{table}

\begin{table*}[tp]
\caption{The knowledge tracing performance obtained with different models on each student subgroup grouped by age. $<$ 80, 80-90, $>$90 represent year of birth prior to 1980, between 1980 and 1990, and after 1990, respectively.}

\begin{center}

\scalebox{0.8}{

\begin{tabular}{cccccccccc}
\hline
\multicolumn{1}{c|}{Dataset}& \multicolumn{3}{c|}{FOC} & \multicolumn{3}{c|}{QDS}  & \multicolumn{3}{c}{MOSFETs}\\  \hline
\multicolumn{1}{c|}{Model} &  $<80$   & $80-90$ & \multicolumn{1}{c|}{$>90$} & $<80$   & $80-90$ & \multicolumn{1}{c|}{$>90$} & $<80$   & $80-90$ & $>90$ \\ \hline\hline
\texttt{[sc2-L]} &  .539 (.011) & .527 (.008) & .535 (.006) & .528 (.007) & .536 (.005) & .523 (.008) & .536 (.010) & .548 (.011) & .531 (.010)\\\hline
\texttt{[sc2-G]} & .553 (.007) & .539 (.005) & .548 (.009) & .536 (.007) & .547 (.010) & .551 (.011) & .568 (.011) & .543 (.008) & .556 (.014)
\\
\texttt{[sc2-G-AV-M]} & .554 (.008) & .546 (.007) & .530 (.008) & .541 (.010) & .559 (.010) & .537 (.003) & .560 (.008) & .546 (.007) & .553 (.011) \\
\texttt{[sc2-G-AT-M]} & .562 (.015) & .557 (.017) & .549 (.008) & .558 (.009) & .567 (.013) & .564 (.015) & .572 (.018) & .568 (.015) & .570 (.019)  \\
\texttt{[sc2-G-AV-T]} & .532 (.008) & .546 (.011) & .538 (.009) & .557 (.009) & .552 (.011) & .546 (.013) & .553 (.015) & .532 (.012) & .548 (.013)  \\
\texttt{[sc2-G-AT-T]} & .546 (.015) & .563 (.018) & .551 (.012) & .562 (.009) & .549 (.013) & .538 (.008) & .546 (.008) & .558 (.015) & .563 (.019)  \\
\hline
\texttt{FedIRT} & .610 (.015) & .639 (.017)& .614 (.015)& .631 (.013)& .600 (.011) & .617 (.015) & .608 (.017)& .592 (.009)& .599 (.007)\\
\texttt{[sc2-P-AV-M]} & .638 (.023) & .652 (.018) & .661 (.018) & .630 (.021) & .643 (.024) & .650 (.019) & .638 (.017) & .620 (.019) & .625 (.021)   \\
\texttt{[sc2-P-AT-M]}(MLPFL) & .654 (.017) & .661 (.015) & .673 (.022) & .647 (.019) & .632 (.018) & .656 (.023) & .640 (.015) & .631 (.021) & .617 (.024)  \\
\texttt{[sc2-P-AV-B]} & .672 (.029) & .669 (.025) & .680 (.028) & \textbf{.660 (.025)} & .672 (.019) & .648 (.020) & .631 (.017) & .650 (.021) & .628 (.020)  \\
\texttt{[sc2-P-AT-B]}(MLPFL)& \textbf{.689 (.025)} & \textbf{.672 (.022)} & \textbf{.699 (.018)} & .654 (.015) & \textbf{.683 (.020)} & \textbf{.661 (.027)} & \textbf{.648 (.022)} & \textbf{.657 (.017)} & \textbf{.631 (.019)}  \\
\hline

\end{tabular}
}
\end{center}
% \vspace{-0.1in}
\vspace{-0.2in}

\label{tb:sc2_kt_a1}
\end{table*}

\begin{table*}[htp]
\caption{Performance variation (standard deviation) across the gender (\texttt{G}), continent (\texttt{C}), and year of birth (\texttt{Y}) demographic variables. The standard deviation is reported as a percentage of the mean.
 \color{black}}

\begin{center}

\scalebox{0.8}{

\begin{tabular}{lccccccccccccccccccc}
\hline
\multicolumn{1}{c|}{Task}& \multicolumn{9}{c|}{Knowledge Tracing} & \multicolumn{9}{c}{Outcome Prediction} \\  \hline
\multicolumn{1}{c|}{Dataset}& \multicolumn{3}{c|}{FOC} & \multicolumn{3}{c|}{QDS}  & \multicolumn{3}{c|}{MOSFETs} & \multicolumn{3}{c|}{FOC} & \multicolumn{3}{c|}{QDS}  & \multicolumn{3}{c}{MOSFETs}\\  \hline
\multicolumn{1}{c|}{Model} & \texttt{G} & \texttt{C} & \multicolumn{1}{c|}{\texttt{Y}} & \texttt{G} & \texttt{C} & \multicolumn{1}{c|}{\texttt{Y}} & \texttt{G} & \texttt{C} & \multicolumn{1}{c|}{\texttt{Y}} & \texttt{G} & \texttt{C} & \multicolumn{1}{c|}{\texttt{Y}} & \texttt{G} & \texttt{C} & \multicolumn{1}{c|}{\texttt{Y}}  & \texttt{G} & \texttt{C} & \multicolumn{1}{c}{\texttt{Y}} \\
\hline

\texttt{[sc1-G]} & 5.29 & 5.64 & 4.52& 8.45&4.23 &5.34 & 4.56  & 3.34 & 3.85 & 7.39 & 4.56  & 3.45 & 9.28 & 3.95 & 4.98 & 5.67 & 4.44 & 3.22\\
\texttt{[sc1-P-AT]}(MLPFL)&  7.62 & 4.97 & 5.67 & 9.25 & 6.66 & 5.99 & 7.22 & 3.49 & 4.58 & 9.40 & 3.96 & 3.95 & 13.2 & 4.52 & 3.88 & 6.85 & 5.42 & 4.98\\ \hline
\texttt{[sc2-P-AT-M]}(MLPFL)& 3.32 & 1.07 & \textbf{0.97} & 0.92 & 1.15 & \textbf{1.21} & 1.20 & 2.04 & \textbf{1.16} & 1.77 & \textbf{1.77} & \textbf{0.76} & \textbf{1.20}& 2.16 & 1.41 & 3.11& 1.82& 1.30\\
\texttt{[sc2-P-AT-B]}(MLPFL)& \textbf{2.68} & \textbf{0.89} & 1.55 & \textbf{0.92} & \textbf{1.15} & 1.51 & \textbf{0.92} & \textbf{1.29} & 1.32 & \textbf{0.21} & 1.84 & 1.20 & 1.48 & \textbf{1.54} & \textbf{1.08} & \textbf{2.40} & \textbf{1.23} & \textbf{1.15}\\
\hline

\end{tabular}
}
\end{center}
% \vspace{-0.1in}
% \vspace{-0.3in}
\vspace{-0.2in}

\label{tb:sc2_improve}
\end{table*}

\vspace{-0.2in}
\subsection{Embedding Visualization}
\label{sec:tsne}
% We employ t-SNE to visualize the learned students' activity representations in a 2D space to qualitatively assess the intepretability of our methodology.
We utilize t-SNE to visualize the acquired student activity representations in a 2D space, aiming to qualitatively evaluate our approach.
t-SNE (t-distributed stochastic neighbor embedding) is a dimensionality reduction technique used to visualize high-dimensional data in two or three dimensions. With t-SNE, datapoints that are statistically close together (i.e., similar) in the original space will be close to one another in the mapped space with high probability, while those far apart (i.e., dissimilar) in the original space will have a low probability of being close in the mapped space. Thus, t-SNE helps reveal similarity patterns in a high-dimensional dataset, such as clusters. For our purposes, a method’s embeddings can be considered “better” if its t-SNE visualizations have more discernible clusters, with a high tendency of the points in each cluster to be from a specific course (in Scenario I) or course-demographic pair (in Scenario II): this indicates that the embeddings contain activity patterns that are important to modeling the differences between users in different courses and demographics for personalized analytics.
\color{black}

\begin{figure*}[t]
    \centering
    \setlength{\abovecaptionskip}{1mm}
    \includegraphics[width=0.85\linewidth]{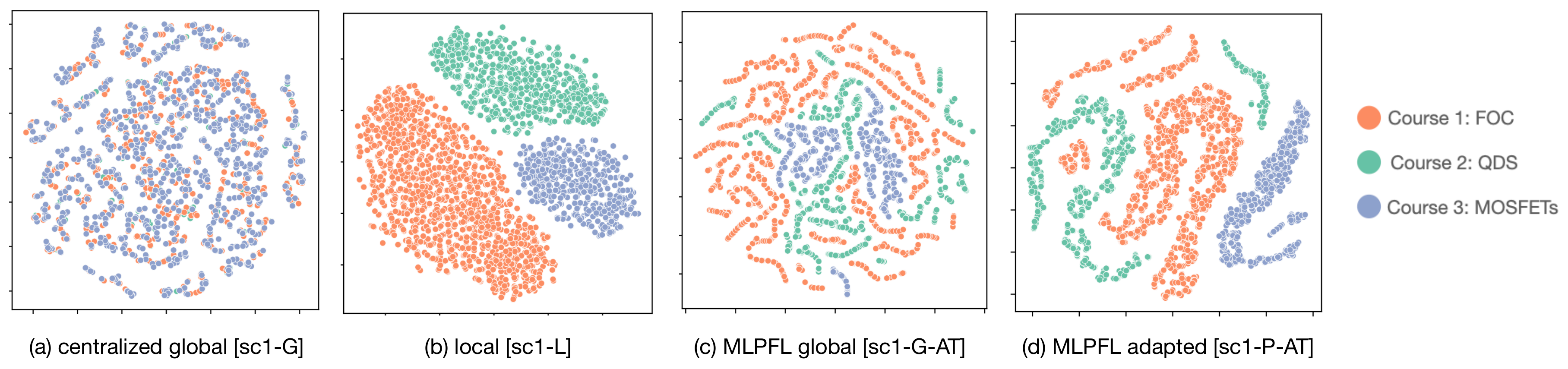}
	 \caption{The embedding vectors of students' knowledge state learned by the centralized method (\texttt{[sc1-G]}), local model (\texttt{[sc1-L]}), federated global model of MLPFL (\texttt{[sc1-G-AT]}), and the personalized model of MLPFL (\texttt{[sc1-P-AT]}) according to different courses in scenario I. 
  % The organized and even clustered representation vectors learned by {\ourmethod} are consistent with its design to learn unique representations for each course.
}
  \vspace{-0.15in}

	  \label{fig:sc1_visual}
\end{figure*}

\begin{figure*}[t]
    \centering
    \setlength{\abovecaptionskip}{1mm}
    \includegraphics[width=0.85\linewidth]{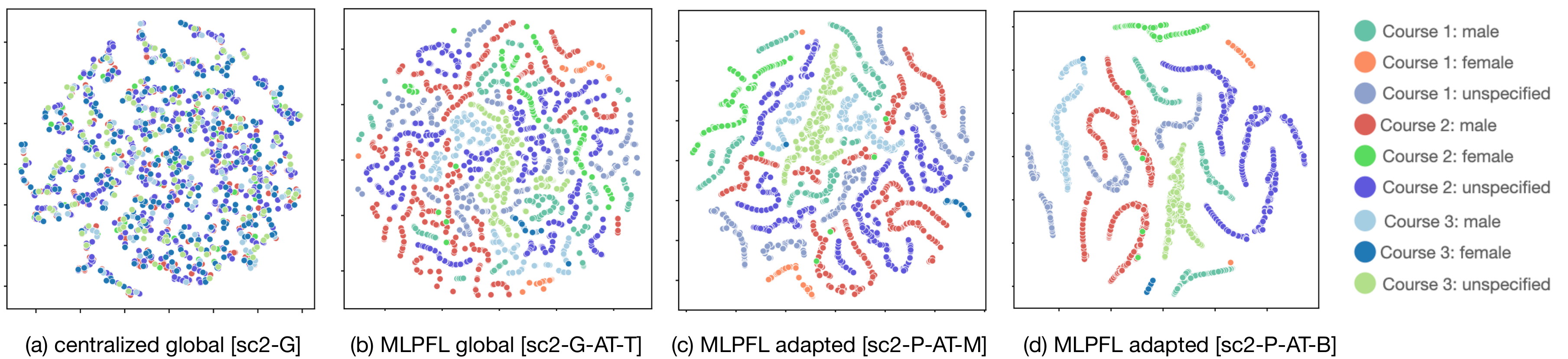}
	 \caption{{Students' knowledge state representations from the centralized method (\texttt{[sc2-G]}) and different hierarchies of our MLPFL in scenario II.} \color{black}
  The organized and even clustered patterns learned by \texttt{[sc2-P-AT-B]} are consistent with its design to learn unique representations for each subgroup.}
  \vspace{-0.15in}
\label{fig:sc2_visual_1}
\end{figure*}

In Figure~\ref{fig:sc1_visual}, we plot the student embeddings for knowledge tracing according to different courses in scenario I, while Figure~\ref{fig:sc2_visual_1} shows the embeddings based on the gender demographic grouping in scenario II.
We defer the visualizations for other demographic subgroups in scenario II to Appendix~\ref{sec:appendix-b}; the results are qualitatively similar.
% All models are trained on $\Omega^{\mathsf{train}}$, and the student embeddings are the combined hidden state $\widetilde{\textbf{h}}_u$ (Section ~\ref{sec:outcomeprediction}) after the attention module that is used to predict their learning outcomes.
All models are trained using $\Omega^{\mathsf{train}}$, and student embeddings are derived from the hidden state $\widetilde{\textbf{h}}_u$ (Section ~\ref{sec:outcomeprediction}), which is obtained after applying the attention module to predict their learning outcomes.
The colors of the dots represent the corresponding student groups.

In Figure~\ref{fig:sc1_visual}(a), the centralized global model does not generate embeddings that differentiate students in different courses for scenario I.
In Figure~\ref{fig:sc1_visual}(b), unsurprisingly, local modeling produces embeddings which cluster by course with linear separation between the clusters. 
Compared to these, the embeddings learned by FL (Figure~\ref{fig:sc1_visual}(c)\&(d)) also demonstrate clustering patterns according to different courses.
In Figure~\ref{fig:sc1_visual}(d), the distribution learned by MLPFL is more separated by course due to the meta-learning adaptation procedure applied to the model in Figure~\ref{fig:sc1_visual}(c).
%This clustering pattern is consistent with our method preserving local attributes pertaining to each course while leveraging global commonalities in student modeling that are present across courses.

For scenario II, in Figure~\ref{fig:sc2_visual_1}, it is hard to differentiate different courses and subgroups from the centralized global model (Figure~\ref{fig:sc2_visual_1}(a)).
Starting with the federated global model (Figure~\ref{fig:sc2_visual_1}(b)), a more clustered distribution emerges as personalization steps are taken to the middle and bottom of the hierarchy in  Figure~\ref{fig:sc2_visual_1}(c) and~\ref{fig:sc2_visual_1}(d).
Compared to the course-level personalized model in Figure~\ref{fig:sc2_visual_1}(c), the MLPFL model personalized by demographic information in Figure~\ref{fig:sc2_visual_1}(d) shows even more well-separated clusters due to additional meta-updates. The same clustering pattern when personalizing according to different continent and age information can be seen from Figure~\ref{fig:sc2_visual_2} in Section~\ref{sec:appendix} of the Appendix.
% These visualizations confirm the ability of {\ourmethod} to adapt a global model to each student subgroup, helping us learn more expressive and predictive representations of student behavior, since we exploit available information unique to each subgroup that would be otherwise discarded.
%These visualizations demonstrate MLPFL's ability to tailor a global model to each student subgroup, enhancing the predictive representations of student behavior by leveraging unique subgroup-specific information.

% These visuals show that student embeddings are at least moderately correlated with the specific student subgroup.
% For instance, in case of gender (Figure~\ref{fig:sc2_visual_1}(c)), males in course 3 tend to appear on the left side of the figure, females in course 2 tend to be at the top, and females in course 1 are on the upper right of the figure. 
Moreover, these visualizations indicate a reasonable level of correlation between activity embeddings and the respective courses and demographic subgroups. For example, considering gender (Figure~\ref{fig:sc2_visual_1}(c)), males from course 3 tend to cluster on the left side, while females from course 2 are often positioned towards the top, and females from course 1 are located on the upper right part of the figure.
% The same phenomenon can be observed in other subplots of Figure~\ref{fig:sc2_visual_2}(a3)\&(b3).
Similar patterns are also evident in the remaining subplots of Figure~\ref{fig:sc2_visual_2}(a3)\&(b3) in the Appendix.
\textit{These results confirm that distinct learning behaviors exist among various student subgroups, which our meta-learning-based personalization method addresses by adapting global models to accommodate these differences.}
\vspace{-0.1in}

% \begin{figure*}[t]
%     \centering
%     \setlength{\abovecaptionskip}{1mm}
%     \includegraphics[width=\linewidth]{figures/sc2_other.pdf}
% 	 \caption{Students' knowledge state representations learned by  different models from our {\ourmethod} training pipeline according to (a1-3) continent and (b1-3) age information. The models include federated global model of {\ourmethod} \texttt{[sc2-G-AT-T]}, {\ourmethod} personalized by courses (\texttt{[sc2-P-AT-M]}), and {\ourmethod} personalized by demographic subgroups (\texttt{[sc2-P-AT-B]}). The organized and even clustered representation vectors learned by our method are consistent with its design to learn unique representations for each subgroup.
% } \label{fig:sc2_visual_2}
% \end{figure*}
\vspace{-0.1in}
\section{Discussion on Real-World Implementation \\ and Applications}
We now discuss how our MLPFL methodology can be integrated into online education platforms, such as the edX platform which hosted the courses we evaluated on in Section~\ref{sec:exp}, to mitigate biases in predictive learning analytics. The most direct application is to integrate MLPFL into the learning management system (LMS) backends, and provide the prediction outputs to students and instructors on their corresponding dashboards. For the outcome prediction and knowledge tracing tasks evaluated in our work, this would include inferences of which students are expected to answer specific in-video quiz questions incorrectly (from knowledge tracing), and which students are at risk of not passing the course (from outcome prediction). Based on this information, instructors could plan targeted interventions to help individual students (or a group of students) in need, e.g., as discussed in~\cite{Chen2019EarlyDP}. We envision these predictions to be updated regularly as students generate more activities in their courses, and to include a measure of confidence in the prediction, derived from the performance indicators observed during the model training procedure (e.g., the AUC values displayed in our analysis).

To evaluate the effectiveness of MLPFL-enhanced dashboards, it will be important to conduct proper A/B testing. We envision establishing three groups for comparison: (A) courses without predictive analytics in dashboards, (B) courses with baseline predictive learning analytics solutions (e.g., {\tt FedIRT}, {\tt [sc2-L]}, {\tt [sc2-G]} from Section~\ref{sec:exp}) integrated into dashboards, and (B) courses with MLPFL integrated in dashboards for mitigating biases. After running such an experiment, one can compare the resulting distributions of content engagement, exam grades, pass/fail results, and other relevant metrics between these groups, with the aim of showing that group (C) has the most successful students, and importantly, the lowest variance across demographic subgroups. One can also quantify the impact of interventions staged by the instructor in each group by comparing student performance before and after such interventions took place. Additionally, qualitative feedback from students and instructors can offer valuable insights into how the dashboard analytics impacted the course experience.

Another application of MLPFL that we envision is to mitigate biases in AI-enabled adaptive learning systems~\cite{Okubo2023AdaptiveLS, Brinton2015IndividualizationFE} that have been integrated into online education platforms. Many of these systems rely on identifying students at risk of early dropout and/or poor grades to determine refinements of their suggested learning paths~\cite{Eegdeman2022ComputerOT, Brinton2015IndividualizationFE}, including but not limited to providing additional resources on particular topics and personalized quiz difficulties. The aim of MLPFL in these systems will be to improve and reduce the variance in quality of these identifications across demographic subgroups. The evaluation of MLPFL here should similarly follow proper A/B testing procedures, where the learning outcomes of an adaptive learning system with and without the integration of our solution are compared.

% Introducing our MLPFL methodology into online education platforms like edX offers potential for tailoring student learning experiences and addressing biases. By leveraging MLPFL, educational platforms can dynamically modify learning content and assessments to better suit the diverse student population, ensuring a more inclusive learning environment. For instance, the implementation of adaptive learning modules that adjust in real-time to a student's interactions with course material is a direct application of MLPFL. Moreover, MLPFL can enhance instructor dashboards with insights for early prediction of students at risk, enabling timely recommendation. This includes providing additional resources for struggling topics, driven by the analytics MLPFL generates.

% The personalized learning paths created through MLPFL specifically account for each student's academic performance, learning preferences, and backgrounds, benefiting underrepresented groups significantly. Traditional educational systems often neglect these factors, but MLPFL’s focus on individual students ensures that underrepresented groups receive tailored support, potentially closing achievement gaps. To assess MLPFL's effectiveness in online courses, we use metrics like engagement rates, completion rates, and performance assessments (e.g., grades). These measures, combined with student feedback on their learning experience, form a comprehensive framework for evaluating MLPFL's impact.

\vspace{-0.1in}
\section{Discussion on Limitations}
We finally discuss some limitations of our study which can motivate future work. One limitation is its dependency on the availability of demographic data, which students may not always be willing to provide. Even in the three courses we studied, a substantial portion of students chose not to specify their gender or age (see Table~\ref{tb:combined_demo}). Such limitations can affect MLPFL’s ability to accurately personalize learning analytics for all students; indeed, we saw in Table~\ref{tb:sc2_improve} that a key disadvantage of scenario I (which does not employ demographic information) is that it can actually increase performance biases slightly. Nonetheless, our results have shown that even when a significant amount of demographic data is missing, MLPFL in scenario II is able to obtain substantial improvements in prediction quality and performance variance across student subgroups. Related to this, another limitation of our study is that our grouping of demographic variables was largely based on intuition from the available data, and may not be appropriate for all course populations. In particular, we modeled the geographic variable based on continent since many countries did not have more than a few samples, and we chose three groups for the age variable somewhat arbitrarily. Future work could investigate a technique for rigorously optimizing the partitioning of a subgroup variable according to performance and bias mitigation objectives.
\color{black}

\vspace{-0.1in}

\section{Conclusion}

In this paper, we developed a Multi-Layer Personalized Federated Learning (MLPFL) framework for mitigating biases in student modeling stemming from data availability. 
Our approach is based on meta learning, adapting local models for various student subgroups from a common global model designed to promote personalization capability.
%\yun{pretraining contribution here?}
%Moreover, we develop a personalized federated learning framework where local models of different student subgroups are adapted from the global model. 
%The adaptation with the meta-learning approach personalize local models based on the data that is localize to the specific student subgroup instead of using data quality heuristics.
We applied our framework to two student modeling tasks: (i) knowledge tracing and (ii) outcome prediction, modeling student activities when they interact with an online learning platform.
Our proposed method considers personalization in a hierarchical manner: first by course (if there are multiple courses), and then by student demographic subgroup within each course (if this information is provided to the model).
Evaluation on three online course datasets showed that our approach surpasses baseline methods by enhancing prediction accuracy across courses and student subgroups, both in terms of mean and variance. 
Moreover, well-organized student embeddings learned by our method were seen to be correlated with improved student modeling. 
% Avenues for future work include (i) exploring more rigorous student subgroup definitions \cite{kearns} and (ii) applying our framework to other types of student models.

\vspace{-0.1in}
\section*{Acknowledgements}
Y. Chu and C. Brinton acknowledge support from the National Science Foundation under grant CNS-2146171.

% \vfill
\bibliographystyle{IEEEtran}
\bibliography{ref-base}

% \begin{IEEEbiography}[{\includegraphics[width=1in,height=1.25in,clip,keepaspectratio]{biofigure/chu_cut.jpg}}]{Yun-Wei Chu}{\space} is currently working toward the PhD degree with the Department of Electrical and Computer Engineering, Purdue University, USA. His current research interests include federated learning and meta-learning.
% \end{IEEEbiography}

\begin{IEEEbiographynophoto}{Yun-Wei Chu}
is currently working toward the PhD degree with the Department of Electrical and Computer Engineering, Purdue University, IN, USA. He received his M.S. degree in Electrical and Control Engineering from National Chiao Tung University in 2017. His current research interests span areas including Natural Language Processing, Machine Learning, and AI for social good.
\end{IEEEbiographynophoto}

\begin{IEEEbiographynophoto}{Seyyedali Hosseinalipour} (S’17, M’20) received the Ph.D. degree in Electrical Engineering from North Carolina State University, NC, USA 2020. He was the recipient of the ECE Doctoral Scholar of the Year Award (2020) and ECE Distinguished Dissertation Award (2021) at North Carolina State University. He has served as the TPC Co-Chair of workshops related to federated learning and fog computing at several conferences such as IEEE INFOCOM, IEEE GLOBECOM, IEEE ICC, and IEEE MSN.  He is currently an Assistant Professor of Electrical Engineering Department at University at Buffalo—SUNY. His research interests include the analysis of modern wireless networks and communication systems, distributed machine learning, and network optimization.
\end{IEEEbiographynophoto}

\begin{IEEEbiographynophoto}{Elizabeth Tenorio} is a data engineer with expertise in building data-driven applications across diverse domains including education, health, cybersecurity, and robotics. In her past roles as software engineer and data scientist, she developed apps for evidence-based cancer care at Vermonster, a content topic analysis platform at Zoomi AI, a drug discovery database at PMS-ICBG, and cyber behavior analytics tools at Forcepoint X-Labs. She serves as a data consultant for Purdue University and a senior data engineer at iRobot. Dr. Tenorio received her PhD in Biology from Kobe University and completed research fellowships in microbiology and bioinformatics at UCLA and Tufts Medical Center.
\end{IEEEbiographynophoto}

\begin{IEEEbiographynophoto}{Laura Melissa Cruz Castro} is an instructional assistant professor at the University of Florida. Her research interests focus on computational thinking at scale and data science education. She looks at K-12, higher education, professional, and community settings for both research interests. She holds a bachelor’s degree in statistics from Universidad Nacional de Colombia, an M.S. in computer engineering, and Ph.D. in engineering education from Purdue University. 
\end{IEEEbiographynophoto}

\begin{IEEEbiographynophoto}{Kerrie A. Douglas}(Member, IEEE) received the Ph.D. degree in  educational studies with concentration on measurement and evaluation from Purdue University, West Lafayette, IN, USA, in 2012. She is an Associate Professor of Engineering Education with Purdue University and Deputy-Director of a large US Department of Defense- funded workforce development consortium for microelectronics, Scalable Asymmetric Lifecycle Engagement (SCALE). Dr. Douglas is a 2021 US NSF CAREER award recipient for her research on increasing the fairness of assessments in engineering education. 
\end{IEEEbiographynophoto}

\begin{IEEEbiographynophoto}{Andrew (Shiting) Lan} is an assistant professor in the Manning College of Information and Computer Sciences, University of Massachusetts Amherst. His research focuses on the development of artificial intelligence (AI) and natural language processing (NLP) methods to enable scalable and effective personalized learning in education. He has worked on many research topics in this domain, including learner modeling, personalization policy learning, automated question and feedback generation, and educational process data analysis. His work resulted in several top prizes at public educational data mining challenges, including ones organized by the US Department of Education.
\end{IEEEbiographynophoto}

\begin{IEEEbiographynophoto}{Christopher G. Brinton} (S’08, M’16, SM’20) is the Elmore Rising Star Assistant Professor of Electrical and Computer Engineering at Purdue University. His research interest is at the intersection of networked systems and machine learning, specifically in distributed machine learning, fog/edge network intelligence, and data-driven network optimization. Dr. Brinton is a recipient of the NSF CAREER Award, ONR Young Investigator Program (YIP) Award, DARPA Young Faculty Award (YFA), AFOSR YIP Award, and Intel Rising Star Faculty Award. He currently serves as an Associate Editor for IEEE/ACM Transactions on Networking. Prior to joining Purdue, Dr. Brinton was the Associate Director of the EDGE Lab and a Lecturer of Electrical Engineering at Princeton University. Dr. Brinton received the PhD (with honors) and MS Degrees from Princeton in 2016 and 2013, respectively, both in Electrical Engineering.
\end{IEEEbiographynophoto}

\appendices
\clearpage
\section{Additional Course and Dataset Details}
\label{sec:data-dist}
This section provides supplementary information to the course descriptions in Section~\ref{sec:data-task}. 

Table~\ref{tb:combined_demo} shows the population and pass rate distributions for each course and student subgroup.
Our research utilized datasets from three courses hosted by Purdue University Online between September 2020 and September 2021: Fiber Optic Communications (FOC), Quantum Detectors and Sensors (QDS), and Essentials of MOSFETs (MOSFETs). Each of these are graduate-level courses from the Electrical and Computer Engineering (ECE) discipline, with different instructors. All three courses followed a consistent format that included a series of lecture videos with end-of-video quizzes (exact numbers given in Table I) accompanied by forums for student discussions. Each course was delivered in English with video transcripts and follows an instructor-paced format with periodic content releases and fixed due dates. Upon completion, students who passed the course were eligible to receive a certificate. Course durations varied: QDS was 17 weeks, FOC was 5 weeks, and MOSFETs was 6 weeks. Each course had two offerings over this year period, which were combined into a single dataset. Targeting graduate-level students, these courses have specific prerequisites: FOC requires knowledge of electric and magnetic fields; QDS expects understanding of electromagnetic fields and basic differential equations; MOSFETs demands a background in undergraduate physics, chemistry, and mathematics, including differential equations

We construct student activity sequences according to the consistent structure of engagement with lecture videos, interactions within the course forums, and responses to in-video quiz questions contained in these three courses, as detailed in Section III-B. These data types are commonly used in student modeling research and are standard across many online education platforms. We acknowledge that the edX platform offers additional functionalities, such as the Notes tool, which allows learners to highlight and make notes directly in the course content, and drag-and-drop events, where learners interact with items in a more dynamic fashion. While these features could offer a more comprehensive view of student activities, they were not included in our analysis.

All students from each course were included in our datasets, so long as they had at least one video interaction or forum participation activity recorded. Our research was ideated after the courses had concluded, and thus students were not aware of the existence of our study while they were taking the courses. We received approval from Purdue’s Institutional Review Board (IRB) prior to conducting our research, and adhered to the IRB guidelines, with all student data being anonymized prior to accessing it.

\color{black}

\section{Implementation and Evaluation Metric}
\label{sec:appendix-implementation}
We employ 5-fold cross-validation on each dataset, randomly selecting 20\% of students from the training set for validation. All neural network architectures, including LSTM for Knowledge Tracing and GRU for Outcome Prediction, use a hidden state dimension of 48.
We employ 50 training epochs for the non-FL approaches (\texttt{[sc1-L]}, \texttt{[sc1-G]}, \texttt{[sc2-L]}, and \texttt{[sc2-G]}), while for FL methods, the parameters for global aggregations ($K$) and local iterations ($E$) are set at 10 and 5, respectively. 
Hyperparameters for all models are determined through a grid search within the parameter space specified in Table~\ref{tb:parameter_main}. 
We evaluate all parameter combinations for each model and select the one yielding the highest accuracy in Section ~\ref{sec:result}.

\vspace{-0.1in}
\begin{table}[!ht]
\caption{Hyperparameters space, with the best-performing combination highlighted in boldface for our method.}

\begin{center}
\scalebox{1}{
\begin{tabular}{c|l}
\hline
Hyper-parameter & Search Space \\\hline
Learning Rate & 1e-2, 5e-2, \textbf{1e-3}, 5e-3, 1e-4, 1e-5, 1e-6\\
Learning Rate Decay & 5e-2 , \textbf{1e-3}, 5e-3, 1e-4\\
Dropout Rate & 0.3, 0.4, \textbf{0.5}, 0.6, 0.7\\
Optimizer & \textbf{Adam}, SGD \\
Batch Size & 4, \textbf{8}, 16, 32\\

\hline
\end{tabular}}
\end{center}
% \vspace{-3mm}
\label{tb:parameter_main}
\vspace{-3mm}
\end{table}

\begin{table*}[!ht]
\caption{
(Left) Population distribution and (Right) pass rates for each student subgroup in every course.  $\Omega^I$ and $\Omega^{\neg I}$ represent students, considering whether they answer questions on demographic variables $I \in \{\texttt{G, C, Y}\}$ when registering for the course, where \texttt{G}, \texttt{C}, and \texttt{Y} represent gender, continent, and year of birth, respectively.}

\begin{center}
\scalebox{0.85}{
\begin{tabular}{l||ccclccc}
\cline{1-4} \cline{6-8}  Dataset & FOC & QDS & MOSFETs &&  FOC & QDS & MOSFETs
\\ \cline{1-4} \cline{6-8}
Overall Statistics & 1,265 & 2,304 & 886 & & 29.3\% & 24.8\% & 28.5\% \\
\hline \hline
 \multicolumn{8}{c}{Gender (\texttt{G})}\\
\hline
Male ($\Omega^{\texttt{G,M}}$) & 42.7\% & 42.9\%  & 43.2\% & & 27.7\% & 31.8\% & 33.6\%\\ 
 Female ($\Omega^{\texttt{G,F}}$)& 8.3\% & 11.3\% & 23.7\% &  & 23.0\% & 32.5\% & 30.9\%\\
Unspecified ($\Omega^{\neg \texttt{G}}$)& 49.0\% & 45.8\% & 33.1\%
& & 27.2\% & 31.8\% & 30.1\%\\
\hline
\multicolumn{8}{c}{Continent (\texttt{C})}\\
\hline
Asian ($\Omega^{\texttt{C,AS}}$) & 35.0\% & 42.8\%  & 41.4\% &  & 31.4\% & 34.5\% & 36.7\%\\ 
African ($\Omega^{\texttt{C,AF}}$) & 10.1\% & 6.9\% & 7.2\% & &  8.5\% & 5.2\% & 10.6\%\\
European ($\Omega^{\texttt{C,EU}}$) & 14.4\% & 13.3\% & 14.6\% & &11.2\% & 9.9\% & 17.7\%\\
North American ($\Omega^{\texttt{C,NA}}$)& 24.5\% & 21.9\% & 21.9\% & &19.8\% & 18.3\% & 28.1\%\\
South American ($\Omega^{\texttt{C,SA}}$) & 5.5\% & 4.2\% & 3.9\% & & 4.4\% & 2.8\% & 10.0\%\\
Unspecified ($\Omega^{\neg \texttt{C}}$) & 10.5\% & 10.9\% & 11.0\%& & 4.6\% & 6.7\% & 11.4\%\\
\hline
\multicolumn{8}{c}{Year of Birth (\texttt{Y})}\\
\hline
Y $\leq$ 1980 ($\Omega^{\texttt{Y, <80}}$) & 12.1\% & 9.0\% & 14.0\%& & 18.7\% & 29.5\% & 27.9\%\\
1980 $<$ Y $\leq$ 1990 ($\Omega^{\texttt{Y, 80-90}}$) & 17.9\% & 16.2\% & 15.2\%& & 26.5\% & 28.2\% & 34.8\%\\
Y $>$ 1990 ($\Omega^{\texttt{Y, >90}}$) & 22.8\% & 29.8\% & 31.5\% & & 32.7\% & 23.7\% & 30.9\%\\
Unspecified ($\Omega^{\neg \texttt{Y}}$) & 42.7\% & 45.0\% & 39.3\%& &28.8\% & 22.1\% & 28.6\%\\
\cline{1-4} \cline{6-8} 

% \vspace{-0.25in}

\end{tabular}}
\end{center}
\label{tb:combined_demo}
\end{table*}

To assess prediction performance, we use the Area Under the ROC Curve (AUC) metric, where random guessing yields an AUC of 0.5 and a perfect model scores 1. We run each experiment five times with different initializations and report average AUC results with standard deviation. Evaluations use the model from the training epoch with the highest validation AUC. For scenario I, AUC is calculated for course models after one local iteration on the global model per course. In scenario II, AUC is measured after two adaptation steps: (i) one meta-update on the global model to form course-specific models, and (ii) one local epoch on these models for each student subgroup.

\vspace{-0.1in}

\section{Experimental Results of Outcome Prediction}
\label{sec:appendix-outcome}
This section supplements Section~\ref{sec:result} with experimental results for each outcome prediction method. Tables \ref{tb:sc2_pp_g1}, \ref{tb:sc2_pp_c1}, and \ref{tb:sc2_pp_a1} display predictions grouped by gender, continent, and age, respectively.

\begin{table*}[tp]
\caption{The outcome prediction performance achieved by various models on individual student subgroups grouped by age.}

\begin{center}

\scalebox{0.85}{

\begin{tabular}{cccccccccc}
\hline
\multicolumn{1}{c|}{Dataset}& \multicolumn{3}{c|}{FOC} & \multicolumn{3}{c|}{QDS}  & \multicolumn{3}{c}{MOSFETs}\\  \hline
\multicolumn{1}{c|}{Model} & $<80$   & $80-90$ & \multicolumn{1}{c|}{$>90$} & $<80$  & $80-90$ & \multicolumn{1}{c|}{$>90$} & $<80$  & $80-90$ & $>90$ \\ \hline\hline
\texttt{[sc2-L]} & .582 (.013) & .600 (.019) & .619 (.017) & .617 (.020) & .588 (.018) & .601 (.023) & .559 (.021) & .593 (.015) & .561 (.012) \\\hline
\texttt{[sc2-G]} & .593 (.017) & .579 (.013) & .591 (.021) & .607 (.018) & .612 (.020) & .595 (.008) & .583 (.013) & .617 (.017) & .590 (.015)
\\
\texttt{[sc2-G-AV-M]} & .579 (.017) & .562 (.013) & .557 (.010) & .581 (.010) & .553 (.018) & .562 (.014) & .540 (.010) & .553 (.009) & .567 (.008) \\
\texttt{[sc2-G-AT-M]} & .584 (.010) & .592 (.015) & .567 (.015) & .593 (.017) & .581 (.011) & .578 (.014) & .567 (.013) & .572 (.018) & .583 (.021)  \\
\texttt{[sc2-G-AV-T]} & .569 (.019) & .554 (.015) & .555 (.009) & .571 (.013) & .562 (.021) & .542 (.017) & .537 (.015) & .530 (.008) & .556 (.010)  \\
\texttt{[sc2-G-AT-T]} & .573 (.017) & .562 (.014) & .551 (.011) & .589 (.020) & .574 (.018) & .563 (.014) & .555 (.009) & .561 (.015) & .579 (.011)  \\
\hline
\texttt{FedIRT} & .697 (.008) &	\textbf{.799 (.013)} &	.713 (.011) &	.654 (.008) &	\textbf{.738 (.012)} &	.689 (.005) &	.609 (.007) &	.623 (.008) &	.654 (.011) \\
\texttt{[sc2-P-AV-M]} & .680 (.019) & .692 (.017) & .671 (.020) & .631 (.017) & .660 (.021) & .657 (.019) & .647 (.016) & .649 (.018) & .651 (.019)   \\
\texttt{[sc2-P-AT-M]}(MLPFL) & .692 (.017) & .705 (.023) & .697 (.021) & .654 (.011) & .682 (.015) & .671 (.017) & .653 (.015) & .642 (.018) & .668 (.020)  \\
\texttt{[sc2-P-AV-B]} & .726 (.025) & .739 (.023) & .732 (.023) & .687 (.018) & .693 (.020) & \textbf{.701 (.025)} & \textbf{.691 (.022)} & .682 (.019) & .678 (.014)  \\
\texttt{[sc2-P-AT-B]}(MLPFL) & \textbf{.738 (.023)} & .762 (.029) & \textbf{.750 (.025)} & \textbf{.703 (.031)} & .718 (.022) & .697 (.019) & .689 (.021) & \textbf{.703 (.025)} & \textbf{.712 (.024)}  \\
\hline

\end{tabular}
}
\end{center}
% \vspace{-0.1in}
\label{tb:sc2_pp_a1}
\end{table*}

\begin{table*}[tp]
\caption{The performance on outcome prediction obtained by different models on each gender subgroup.}

\begin{center}

\scalebox{0.85}{

\begin{tabular}{ccccccc}
\hline
\multicolumn{1}{c|}{Dataset}& \multicolumn{2}{c|}{FOC} & \multicolumn{2}{c|}{QDS}  & \multicolumn{2}{c}{MOSFETs}\\  \hline
\multicolumn{1}{c|}{Model} & male & \multicolumn{1}{c|}{female} & male & \multicolumn{1}{c|}{female} & male & female \\ \hline\hline
\texttt{[sc2-L]} & .546 (.021) & .600 (.017) & .625 (.022) & .612 (.024) & .603 (.008) & .655 (.021) \\\hline
\texttt{[sc2-G]}  & .613 (.018) & .632 (.025) & .621 (.021) & .607 (.016) & .631 (.019) & .592 (.007) 
\\
\texttt{[sc2-G-AV-M]} & .544 (.011) & .562 (.021) & .619 (.025) & .603 (.017) & .621 (.019) & .587 (.015) \\
\texttt{[sc2-G-AT-M]} & .603 (.017) & .649 (.013) & .634 (.025) & .600 (.017) & .629 (.009) & .613 (.011) \\
\texttt{[sc2-G-AV-T]} & .558 (.010) & .543 (.008) & .627 (.019) & .600 (.014) & .619 (.013) & .620 (.018) \\
\texttt{[sc2-G-AT-T]} & .612 (.019) & .637 (.017) & .613 (.022) & .584 (.013) & .627 (.027) & .549 (.019) \\
\hline
\texttt{FedIRT} & .702 (.013) &	.658 (.009) &	.679 (.015) &	.617 (.013) & .596 (.012) &	.614 (.010)\\
\texttt{[sc2-P-AV-M]} & .691 (.028) & .725 (.033) & .700 (.031) & .683 (.027) & .642 (.035) & .623 (.025)  \\
\texttt{[sc2-P-AT-M]}(MLPFL) & .718 (.038) & .743 (.026) & .692 (.035) & .709 (.029) & .663 (.025) & .619 (.026) \\
\texttt{[sc2-P-AV-B]} & .748 (.031) & .765 (.028) & .710 (.035) & .719 (.021) & .650 (.027) & .678 (.025) \\
\texttt{[sc2-P-AT-B]}(MLPFL) & \textbf{.769 (.034)} & \textbf{.772 (.019)} & \textbf{.743 (.028)} & \textbf{.722 (.022)} & \textbf{.658 (.018)} & \textbf{.692 (.020)} \\
\hline

\end{tabular}
}
\end{center}
% \vspace{-0.1in}
\label{tb:sc2_pp_g1}
\end{table*}
\begin{table}
\caption{The performance of various models on predicting student outcomes within each student subgroup categorized by continent.}

\begin{center}

\scalebox{0.75}{

\begin{tabular}{cccccc}
\hline
\multicolumn{1}{c|}{Dataset}& \multicolumn{5}{c}{FOC} \\  \hline
 \multicolumn{1}{c|}{Model} & AS & AF & EU & NA & SA \\ \hline\hline
\texttt{[sc2-L]}  & .573 (.009)& .601 (.011)& .615 (.016) &	.539 (.005) &.607 (.013)	
\\\hline

\texttt{[sc2-G]}& .635 (.015)& .602 (.008)& .596 (.016) &	.628 (.015) &.612 (.017)
\\
\texttt{[sc2-G-AV-M]} & .554 (.006)& .573 (.008)& .582 (.012) &	.610 (.016) &.584 (.014)
\\
\texttt{[sc2-G-AT-M]} & .640 (.021)& .602 (.026)& .589 (.018) &	.613 (.016) &.600 (.014)\\
\texttt{[sc2-G-AV-T]} & .594 (.013)& .553 (.015)& .562 (.020) &	.584 (.017) &.573 (.021)\\
\texttt{[sc2-G-AT-T]} & .632 (.023)& .579 (.015)& .552 (.017) &	.632 (.019) &.594 (.013) \\ \hline
\texttt{FedIRT} & .632 (.008) &	.651 (.010) &		\textbf{.733 (.006)} &		.673 (.008) &		.711 (.009)	\\
\texttt{[sc2-P-AV-M]} & .653 (.023)& .648 (.025)& .665 (.026) &	.641 (.029) &.612 (.017) \\
\texttt{[sc2-P-AT-M]}(MLPFL)  & .672 (.026)& .683 (.030)& .709 (.035) &	.683 (.028) &.662 (.021) \\
\texttt{[sc2-P-AV-B]}  & .687 (.028)& .699 (.026)& .723 (.027) &	.678 (.030) &.709 (.034) \\
\texttt{[sc2-P-AT-B]}(MLPFL) & \textbf{.692 (.026)}& \textbf{.713 (.033)}& .720 (.035) &	\textbf{.697 (.028)} &\textbf{.738 (.029)} \\

\hline\hline
\multicolumn{1}{c|}{Dataset}& \multicolumn{5}{c}{QDS} \\  \hline
 \multicolumn{1}{c|}{Model} & AS & AF & EU & NA & SA \\ \hline\hline
\texttt{[sc2-L]} & .612 (.019)& .647 (.015)& .573 (.016) &	.599 (.014) &.628 (.020)
\\\hline

\texttt{[sc2-G]} & .619 (.028)& .604 (.013)& .586 (.009) &	.621 (.016) &.617 (.018)
\\
\texttt{[sc2-G-AV-M]} & .600 (.013)& .623 (.028)& .610 (.023) &	.584 (.017) &.562 (.013)
\\
\texttt{[sc2-G-AT-M]} & .652 (.023)& .638 (.018)& .626 (.016) &	.619 (.014) &.639 (.020)\\
\texttt{[sc2-G-AV-T]}  & .613 (.017)& .600 (.013)& .546 (.008) &	.559 (.009) &.587 (.015) \\
\texttt{[sc2-G-AT-T]} & .671 (.010)& .623 (.017)& .642 (.016) &	.638 (.013) &.601 (.009)\\ \hline
\texttt{FedIRT}& .779 (.008) &	.719 (.011) &		.682 (.009) &		.709 (.012) &	.682 (.015)\\
\texttt{[sc2-P-AV-M]} & .719 (.021)& .764 (.029)& .738 (.024) &	.692 (.018) &.683 (.023) \\
\texttt{[sc2-P-AT-M]}(MLPFL) & .754 (.027)& .778 (.029)& .742 (.031) &	.719 (.024) &.756 (.035)\\
\texttt{[sc2-P-AV-B]} & .762 (.026)& .774 (.027)& \textbf{.776 (.035)} &	.741 (.024) &.758 (.026)\\
\texttt{[sc2-P-AT-B]}(MLPFL) & \textbf{.782 (.022)}& \textbf{.803 (.034)}& .772 (.031) &	\textbf{.762 (.030)} &\textbf{.773 (.028)} \\

\hline\hline
\multicolumn{1}{c|}{Dataset}& \multicolumn{5}{c}{MOSFETs} \\  \hline
 \multicolumn{1}{c|}{Model} & AS & AF & EU & NA & SA \\ \hline\hline
\texttt{[sc2-L]} & .586 (.016)& .554 (.014)& .603 (.018) &	.606 (.016) &.593 (.009)
\\\hline

\texttt{[sc2-G]} & .624 (.020)& .602 (.019)& .587 (.007) &	.592 (.010) &.623 (.016)
\\
\texttt{[sc2-G-AV-M]} & .603 (.020)& .587 (.017)& .559 (.014) &	.560 (.016) &.570 (.010)
\\
\texttt{[sc2-G-AT-M]} & .613 (.016)& .589 (.009)& .552 (.013) &	.578 (.015) &.612 (.018)\\
\texttt{[sc2-G-AV-T]} & .589 (.016)& .563 (.012)& .539 (.013) &	.541 (.009) &.572 (.010) \\
\texttt{[sc2-G-AT-T]} & .602 (.015)& .562 (.013)& .545 (.017) &	.582 (.016) &.600 (.020) \\ \hline
\texttt{FedIRT} &.631 (.010) &	.653 (.013) &		.637 (.009) &		.612  (.008) &		.584 (.006)\\
\texttt{[sc2-P-AV-M]} & .589 (.016)& .613 (.028)& .628 (.016) &	.596 (.011) &.587 (.013) \\
\texttt{[sc2-P-AT-M]}(MLPFL)  & .626 (.026)& .651 (.023)& .632 (.028) &	.628 (.021) &.600 (.016)\\
\texttt{[sc2-P-AV-B]}  & .652 (.025)& .672 (.023)& .640 (.024) &	.679 (.028) &.640 (.026) \\
\texttt{[sc2-P-AT-B]}(MLPFL) & \textbf{.673 (.030)}& \textbf{.684 (.027)}& \textbf{.662 (.025)} &	\textbf{.679 (.026)} &\textbf{.654 (.022)} \\

\hline
\end{tabular}}
\end{center}
\label{tb:sc2_pp_c1}
\vspace{-9mm}
\end{table}

\section{Additional Prediction Results for scenario II}
\label{sec:appendix}
This section supplements Section~\ref{sec:result} with results for students who did not provide demographic information, treating them as an additional subgroup. For knowledge tracing, Tables~\ref{tb:sc2_kt_g2}, \ref{tb:sc2_kt_c2}, and \ref{tb:sc2_kt_a2} show predictions grouped by gender, continent, and age, respectively. Outcome prediction results for these groupings are in Tables~\ref{tb:sc2_pp_g2}, \ref{tb:sc2_pp_c2}, and \ref{tb:sc2_pp_a2}.

% Figure~\ref{fig:mean_var_noinfo} shows the overall average and variance in prediction quality across subgroups for personalized-based algorithms.
% Aligning with the findings we found in Section~\ref{sec:dis_II}, we see that our method provides the strongest prediction quality results and mitigates the data availability biases.

\section{Additional Embedding Visualization for scenario II}
\label{sec:appendix-b}
\begin{figure*}
    \centering
    \setlength{\abovecaptionskip}{1mm}
    \includegraphics[width=0.85\linewidth]{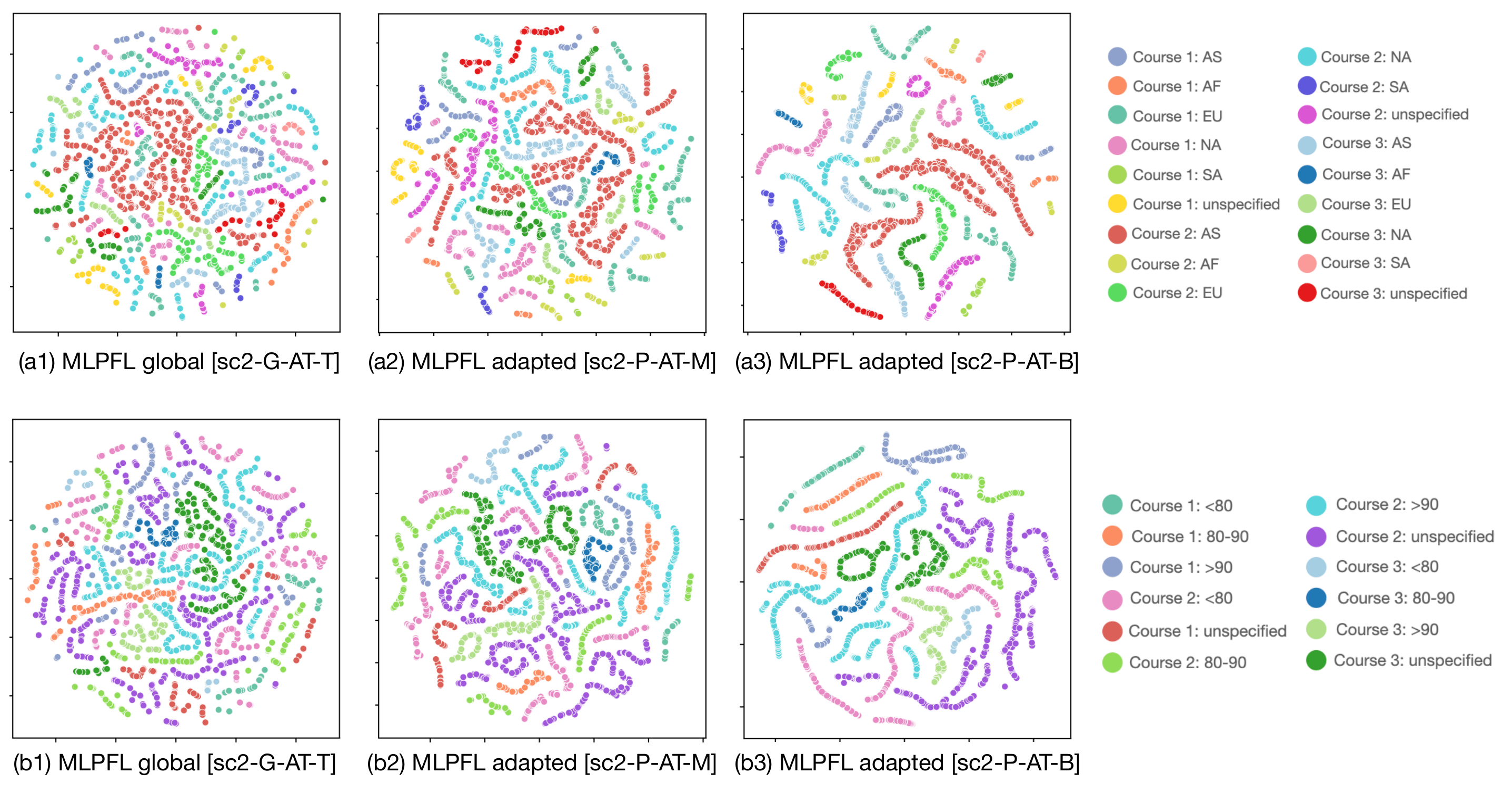}
	 \caption{Students' knowledge state representations learned by  different models from our MLPFL training pipeline according to (a1-3) continent and (b1-3) age information. The models include federated global model of MLPFL \texttt{[sc2-G-AT-T]}, MLPFL personalized by courses (\texttt{[sc2-P-AT-M]}), and MLPFL personalized by demographic subgroups (\texttt{[sc2-P-AT-B]}).}
  \label{fig:sc2_visual_2}
\end{figure*}

In this section, we provide supplemental results to Section~\ref{sec:tsne}.
Figure~\ref{fig:sc2_visual_2} shows the embeddings based on the continent and age demographic grouping in scenario II.
The clustering patterns Figure~\ref{fig:sc2_visual_2} also show that MLPFL learns more expressive and predictive representations of student behavior.

\begin{table}[!ht]
\caption{The knowledge tracing performance achieved by different models on each student subgroup categorized by continent.}

\begin{center}

\scalebox{0.65}{

\begin{tabular}{ccccccc}
\hline
\multicolumn{1}{c|}{Dataset}& \multicolumn{5}{c}{FOC} \\  \hline
 \multicolumn{1}{c|}{Model} & AS & AF & EU & NA & SA & unspecified \\ \hline\hline
\texttt{[sc2-L]} & .521 (.007)& .519 (.005) & .536 (.008)  & .517 (.004)	 & .532 (.009)	& .528 (.005)
\\\hline

\texttt{[sc2-G]} & .556 (.012)& .547 (.008) & .538 (.010)  & .533 (.006)	 & .519 (.002)	& .544 (.007)
\\
\texttt{[sc2-G-AV-M]}  & .547 (.013)& .532 (.009) & .546 (.012)  & .521 (.005)	 & .536 (.010)	& .517 (.003)
\\
\texttt{[sc2-G-AT-M]} & .564 (.013)& .553 (.008) & .567 (.015)  & .532 (.016)	 & .560 (.017)	& .573 (.018)\\
\texttt{[sc2-G-AV-T]} & .539 (.011)& .516 (.004) & .527 (.003)  & .540 (.009)	 & .549 (.015)	& .532 (.015)\\
\texttt{[sc2-G-AT-T]} & .549 (.008)& .538 (.004) & .543 (.013)  & .549 (.010)	 & .554 (.017)	& .557 (.016)\\ \hline
\texttt{FedIRT} & .602 (.013)& .574 (.011)& .592 (.008)& .611 (.011)& .608 (.013)& .589 (.008)\\
\texttt{[sc2-P-AV-M]} & .614 (.019)& .603 (.020) & .600 (.015)  & .617 (.011)	 & .632 (.016)	& .628 (.014) \\
\texttt{[sc2-P-AT-M]}(MLPFL) & .613 (.011)& .627 (.018) & .621 (.014)  & .627 (.015)	 & .632 (.013)	& .614 (.020)\\
\texttt{[sc2-P-AV-B]}  & .632 (.021)& \textbf{.649 (.025)} & .637 (.016)  & .623 (.017)	 & .668 (.026)	& .650 (.017)\\
\texttt{[sc2-P-AT-B]}(MLPFL)& \textbf{.643 (.014)}& .632 (.018) & \textbf{.658 (.016)}  & \textbf{.661 (.015)}	 & \textbf{.673 (.023)}	& \textbf{.662 (.020)}\\

\hline\hline
\multicolumn{1}{c|}{Dataset}& \multicolumn{5}{c}{QDS} \\  \hline
 \multicolumn{1}{c|}{Model} & AS & AF & EU & NA & SA & unspecified\\ \hline\hline
\texttt{[sc2-L]}& .510 (.004)& .533 (.006) & .548 (.008)  & .564 (.010)	 & .523 (.007)	& .527 (.005)
\\\hline

\texttt{[sc2-G]} & .544 (.010)& .543 (.008) & .586 (.016)  & .562 (.013)	 & .533 (.007)	& .547 (.006)
\\
\texttt{[sc2-G-AV-M]} & .538 (.005)& .518 (.003) & .542 (.010)  & .537 (.009)	 & .546 (.006)	& .558 (.004)
\\
\texttt{[sc2-G-AT-M]} & .587 (.010)& .562 (.012) & .574 (.013)  & .568 (.012)	 & .576 (.017)	& .580 (.011)\\
\texttt{[sc2-G-AV-T]} & .558 (.015)& .547 (.013) & .566 (.013)  & .554 (.008)	 & .538 (.004)	& .569 (.010)\\
\texttt{[sc2-G-AT-T]}  & .562 (.012)& .543 (.010) & .560 (.010)  & .573 (.009)	 & .579 (.013)	& .556 (.015)\\ \hline
\texttt{FedIRT}  & .654 (.015)& .617 (.015)& .603 (.008)& .611 (.013)& .586 (.010)& .600 (.009)\\
\texttt{[sc2-P-AV-M]} & .647 (.023)& .648 (.028) & .626 (.020)  & .632 (.018)	 & .619 (.011)	& .594 (.013)\\
\texttt{[sc2-P-AT-M]}(MLPFL)  & .634 (.015)& .658 (.017) & .639 (.021)  & .660 (.021)	 & .651 (.018)	& .615 (.023) \\
\texttt{[sc2-P-AV-B]}  & .654 (.025)& .670 (.029) & .648 (.031)  & .651 (.025)	 & .650 (.024)	& \textbf{.672 (.020)} \\
\texttt{[sc2-P-AT-B]}(MLPFL) & \textbf{.679 (.023)}& \textbf{.698 (.028)} & \textbf{.670 (.019)}  & \textbf{.666 (.022)}	 & \textbf{.669 (.024)}	& .657 (.025)\\

\hline\hline
\multicolumn{1}{c|}{Dataset}& \multicolumn{5}{c}{MOSFETs} \\  \hline
 \multicolumn{1}{c|}{Model} & AS & AF & EU & NA & SA & unspecified\\ \hline\hline
\texttt{[sc2-L]}& .531 (.010)& .518 (.004) & .510 (.006)  & .537 (.008)	 & .555 (.015)	& .521 (.008)
\\\hline

\texttt{[sc2-G]} & .535 (.007)& .567 (.011) & .521 (.006)  & .537 (.005)	 & .561 (.010)	& .537 (.008)
\\
\texttt{[sc2-G-AV-M]} & .521 (.003)& .531 (.005) & .534 (.010)  & .542 (.011)	 & .550 (.008)	& .548 (.013)
\\
\texttt{[sc2-G-AT-M]} & .554 (.015)& .539 (.010) & .557 (.011)  & .563 (.016)	 & .549 (.018)	& .562 (.120)\\
\texttt{[sc2-G-AV-T]} & .542 (.013)& .538 (.015) & .529 (.010)  & .528 (.008)	 & .539 (.013)	& .525 (.017)\\
\texttt{[sc2-G-AT-T]} & .530 (.008)& .517 (.010) & .532 (.009)  & .548 (.014)	 & .560 (.016)	& .554 (.018)\\ \hline
\texttt{FedIRT} & .629 (.015)& .592 (.009)& .573 (.009)& .603 (.010)& \textbf{.618 (.013)}& .608 (.008)\\
\texttt{[sc2-P-AV-M]} & .597 (.016)& .569 (.023) & .554 (.021)  & .596 (.025)	 & .610 (.028)	& .607 (.022) \\
\texttt{[sc2-P-AT-M]}(MLPFL) & .593 (.022)& .580 (.021) & .536 (.010)  & .579 (.015)	 & .592 (.017)	& .614 (.019)\\
\texttt{[sc2-P-AV-B]} & \textbf{.632 (.031)}& .597 (.028) & .576 (.022)  & .616 (.020)	 & .617 (.019)	& .624 (.017)\\
\texttt{[sc2-P-AT-B]}(MLPFL) & .614 (.016)& \textbf{.602 (.020)} & \textbf{.598 (.015)}  & \textbf{.621 (.015)}	 & .610 (.023)	& \textbf{.632 (.021)}\\

\hline
\end{tabular}
}
\end{center}
\label{tb:sc2_kt_c2}
\end{table}
\begin{table}[!ht]
\caption{The predictive performance of different models for student outcome on each student subgroup categorized by continent.}

\begin{center}

\scalebox{0.65}{

\begin{tabular}{ccccccc}
\hline
\multicolumn{1}{c|}{Dataset}& \multicolumn{5}{c}{FOC} \\  \hline
 \multicolumn{1}{c|}{Model} & AS & AF & EU & NA & SA & unspecified \\ \hline\hline
\texttt{[sc2-L]}& .573 (.009)& .601 (.011) & .615 (.016)  & .539 (.005)	 & .607 (.013)	& .548 (.008)
\\\hline

\texttt{[sc2-G]} & .606 (.010)& .587 (.008) & .549 (.005)  & .623 (.011)	 & .610 (.013)	& .584 (.010)
\\
\texttt{[sc2-G-AV-M]} & .559 (.015)& .584 (.010) & .589 (.016)  & .590 (.015)	 & .594 (.014)	& .567 (.016)
\\
\texttt{[sc2-G-AT-M]} & .592 (.009)& .609 (.016) & .601 (.014)  & .613 (.015)	 & .588 (.015)	& .572 (.013)\\
\texttt{[sc2-G-AV-T]} & .593 (.016)& .586 (.017) & .564 (.015)  & .573 (.018)	 & .583 (.020)	& .559 (.014)\\
\texttt{[sc2-G-AT-T]} & .603 (.016)& .572 (.014) & .579 (.015)  & .583 (.009)	 & .591 (.017)	& .553 (.009) \\ \hline
\texttt{FedIRT} & .623 (.009) & \textbf{.702 (.012)} &	 .699 (.014) &		.658 (.011) &		.633 (.015) &		.570 (.016) \\
\texttt{[sc2-P-AV-M]} & .624 (.018)& .650 (.016) & .664 (.018)  & .682 (.016)	 & .698 (.020)	& .682 (.021) \\
\texttt{[sc2-P-AT-M]}(MLPFL)  & .649 (.016)& .654 (.018) & .685 (.023)  & .692 (.021)	 & .684 (.022)	& \textbf{.694 (.021)}\\
 \texttt{[sc2-P-AV-B]}   & .652 (.019)& .683 (.023) & .700 (.027)  & .719 (.025)	 & \textbf{.706 (.022)}	& .674 (.019)\\
 \texttt{[sc2-P-AT-B]}(MLPFL)  & \textbf{.682 (.014)}& .700 (.018) & \textbf{.719 (.023)}  & \textbf{.732 (.022)}	 & .693 (.024)	& .668 (.017)\\

\hline\hline
\multicolumn{1}{c|}{Dataset}& \multicolumn{5}{c}{QDS} \\  \hline
 \multicolumn{1}{c|}{Model} & AS & AF & EU & NA & SA & unspecified\\ \hline\hline
\texttt{[sc2-L]}& .612 (.019)& .647 (.015) & .573 (.016)  & .599 (.014)	 & .628 (.020)	& .532 (.005)
\\\hline

\texttt{[sc2-G]}& .602 (.012)& .615 (.013) & .574 (.011)  & .589 (.015)	 & .613 (.013)	& .570 (.013)
\\
\texttt{[sc2-G-AV-M]} & .563 (.011)& .594 (.013) & .582 (.015)  & .590 (.020)	 & .583 (.013)	& .586 (.014)
\\
\texttt{[sc2-G-AT-M]} & .573 (.015)& .609 (.021) & .592 (.018)  & .584 (.016)	 & .596 (.017)	& .592 (.021)\\
\texttt{[sc2-G-AV-T]} & .589 (.020)& .562 (.017) & .581 (.018)  & .555 (.015)	 & .564 (.011)	& .550 (.016) \\
\texttt{[sc2-G-AT-T]} & .591 (.017)& .572 (.016) & .563 (.015)  & .546 (.010)	 & .580 (.018)	& .562 (.019) \\ \hline
\texttt{FedIRT} & .689 (.011) &		.713 (.017) &		.723 (.013) &		.709 (.018) &		.678 (.009) &		.554 (.014)\\
\texttt{[sc2-P-AV-M]} & .654 (.023)& .680 (.027) & .693 (.022)  & .655 (.021)	 & .651 (.022)	& .667 (.026) \\
\texttt{[sc2-P-AT-M]}(MLPFL)   & .679 (.017)& .701 (.022) & .729 (.028)  & .684 (.015)	 & .673 (.016)	& .684 (.013) \\
 \texttt{[sc2-P-AV-B]}   & .709 (.025)& .743 (.021) & .759 (.022)  & .716 (.019)	 & .700 (.019)	& \textbf{.695 (.017)}\\
 \texttt{[sc2-P-AT-B]}(MLPFL)  & \textbf{.724 (.020)}& \textbf{.770 (.018)} & \textbf{.783 (.023)}  & \textbf{.743 (.025)}	 & \textbf{.712 (.019)}	& .683 (.017) \\

\hline\hline
\multicolumn{1}{c|}{Dataset}& \multicolumn{5}{c}{MOSFETs} \\  \hline
 \multicolumn{1}{c|}{Model} & AS & AF & EU & NA & SA & unspecified\\ \hline\hline
\texttt{[sc2-L]}& .594 (.016)& .553 (.014) & .572 (.018)  & .608 (.016)	 & .553 (.009)	& .586 (.007)
\\\hline

\texttt{[sc2-G]}& .589 (.013)& .572 (.015) & .583 (.016)  & .562 (.014)	 & .566 (.013)	& .555 (.007)	
\\
\texttt{[sc2-G-AV-M]} & .592 (.015)& .554 (.019) & .563 (.016)  & .589 (.014)	 & .564 (.017)	& .549 (.008)
\\
\texttt{[sc2-G-AT-M]} & .601 (.015)& .582 (.014) & .570 (.013)  & .593 (.013)	 & .588 (.018)	& .553 (.015)\\
\texttt{[sc2-G-AV-T]} & .561 (.011)& .523 (.007) & .554 (.009)  & .549 (.010)	 & .570 (.017)	& .581 (.016) \\
\texttt{[sc2-G-AT-T]} & .570 (.015)& .543 (.011) & .585 (.015)  & .573 (.013)	 & .561 (.018)	& .579 (.011) \\ \hline
\texttt{FedIRT} & .663 (.010) &		.683 (.011) &		\textbf{.705 (.015)} &		\textbf{.694 (.011)} &		.688 (.009) &	.692 (.012)\\
\texttt{[sc2-P-AV-M]} & .639 (.019)& .622 (.016) & .600 (.018)  & .641 (.017)	 & .639 (.015)	& .660 (.019) \\
\texttt{[sc2-P-AT-M]}(MLPFL)  & .659 (.011)& .643 (.018) & .613 (.010)  & .633 (.013)	 & .648 (.015)	& .658 (.011)\\
 \texttt{[sc2-P-AV-B]}  & .689 (.020)& .672 (.018) & .676 (.017)  & .661 (.015)	 & .678 (.016)	& .697 (.017)\\
 \texttt{[sc2-P-AT-B]}(MLPFL)  & \textbf{.703 (.023)}& \textbf{.694 (.019)} & .663 (.017)  & .684 (.016)	 & \textbf{.700 (.020)}	& \textbf{.713 (.023)}\\

\hline
\end{tabular}
}
\end{center}
\label{tb:sc2_pp_c2}
\end{table}

\begin{table*}[tp]
\caption{The knowledge tracing performance achieved by various methods for different student subgroups grouped by gender. The ``unspecified” subgroup includes students who did not provide their gender information.}

\begin{center}

\scalebox{0.85}{

\begin{tabular}{cccccccccc}
\hline
\multicolumn{1}{c|}{Dataset}& \multicolumn{3}{c|}{FOC} & \multicolumn{3}{c|}{QDS}  & \multicolumn{3}{c}{MOSFETs}\\  \hline
\multicolumn{1}{c|}{Model} & male & female & \multicolumn{1}{c|}{unspecified} & male & female & \multicolumn{1}{c|}{unspecified} & male & female & unspecified\\ \hline\hline
\texttt{[sc2-L]} & .532 (.009) & .519 (.004) & .523 (.008) & .541 (.010) & .523 (.006) & .531 (.011) & .538 (.011) & .519 (.003) & .517 (.008)\\\hline
\texttt{[sc2-G]}  & .552 (.006) & .541 (.008) & .539 (.006) & .548 (.013) & .537 (.010) & .554 (.006) & .558 (.013) & .541 (.011) & .532 (.010)
\\
\texttt{[sc2-G-AV-M]} & .543 (.019) & .527 (.004) & .511 (.003) & .525 (.008) & .530 (.006) & .527 (.010) & .524 (.006) & .533 (.007) & .529 (.009)\\
\texttt{[sc2-G-AT-M]} & .551 (.009) & .543 (.011) & .510 (.003) & .517 (.004) & .529 (.008) & .534 (.006) & .532 (.005) & .540 (.008) & .538 (.010)\\
\texttt{[sc2-G-AV-T]} & .539 (.010) & .548 (.013) & .529 (.006) & .537 (.007) & .518 (.006) & .523 (.004) & .530 (.008) & .527 (.006) & .547 (.011)\\
\texttt{[sc2-G-AT-T]} & .567 (.008) & .558 (.008) & .527 (.006) & .540 (.007) & .539 (.008) & .534 (.011) & .528 (.009) & .534 (.007) & .550 (.013)\\
\hline
\texttt{FedIRT} & .617 (.013)& .629 (.018)& .601 (.014)& .645 (.011)& .600 (.009)& .631 (.010)& .619 (.008)& .592 (.005)& .604 (.010)\\
\texttt{[sc2-P-AV-M]} & .604 (.013) & .593 (.018) & .633 (.016) & .638 (.011) & .631 (.020) & .626 (.018) & .649 (.016) & .652 (.017) & .643 (.021)\\
\texttt{[sc2-P-AT-M]}(MLPFL) & .610 (.013) & .605 (.015) & .639 (.016) & .658 (.011) & .630 (.015) & .649 (.020) & .654 (.021) & .649 (.019) & .632 (.017)\\
\texttt{[sc2-P-AV-B]} & .643 (.022) & .621 (.017) & .647 (.023) & .676 (.016) & .645 (.017) & .660 (.021) & .668 (.028) & .675 (.027) & .664 (.026)\\
\texttt{[sc2-P-AT-B]}(MLPFL) & \textbf{.652 (.017)} & \textbf{.638 (.019)} & \textbf{.647 (.016)} & \textbf{.691 (.021)} & \textbf{.654 (.020)} & \textbf{.663 (.023)} & \textbf{.687 (.016)} & \textbf{.680 (.019)} & \textbf{.655 (.020)}\\
\hline

\end{tabular}
}
\end{center}
% \vspace{-0.1in}
\label{tb:sc2_kt_g2}
\end{table*}

\begin{table*}[tp]
\caption{The outcome prediction performance achieved by various methods for different student subgroups grouped by gender. The ``unspecified” subgroup includes students who did not provide their gender information.}

\begin{center}

\scalebox{0.85}{

\begin{tabular}{cccccccccc}
\hline
\multicolumn{1}{c|}{Dataset}& \multicolumn{3}{c|}{FOC} & \multicolumn{3}{c|}{QDS}  & \multicolumn{3}{c}{MOSFETs}\\  \hline
\multicolumn{1}{c|}{Model} & male & female & \multicolumn{1}{c|}{unspecified} & male & female & \multicolumn{1}{c|}{unspecified} & male & female & unspecified\\ \hline\hline
\texttt{[sc2-L]} & .546 (.021) & .600 (.017) & .562 (.016) & .625 (.022) & .612 (.024) & .617 (.020) & .603 (.008) & .655 (.021) & .604 (.016)\\\hline
\texttt{[sc2-G]} & .631 (.016) & .589 (.018) & .562 (.007) & .604 (.019) & .617 (.015) & .592 (.016) & .590 (.018) & .594 (.014) & .591 (.016)
\\
\texttt{[sc2-G-AV-M]} & .619 (.020) & .603 (.017) & .587 (.016) & .592 (.018) & .584 (.013) & .580 (.027) & .588 (.028) & .581 (.016) & .612 (.024)\\
\texttt{[sc2-G-AT-M]} & .632 (.028) & .613 (.025) & .580 (.019) & .615 (.018) & .600 (.023) & .587 (.017) & .603 (.016) & .612 (.014) & .590 (.019)\\
\texttt{[sc2-G-AV-T]} & .610 (.022) & .584 (.016) & .590 (.017) & .587 (.020) & .591 (.016) & .572 (.018) & .584 (.008) & .593 (.011) & .602 (.017)\\
\texttt{[sc2-G-AT-T]} & .621 (.020) & .590 (.013) & .583 (.017) & .601 (.022) & .593 (.018) & .590 (.013) & .611 (.012) & .600 (.015) & .582 (.013)\\
\hline
\texttt{FedIRT} & .620 (.016) &	.610 (.013) &	.554 (.009) &	 .609 (.011) &	.649 (.012) &	.594 (.010) &	.652 (.013) &	.693 (.011) &	.664 (.015) \\
\texttt{[sc2-P-AV-M]} & .667 (.022) & .659 (.028) & .631 (.026) & .690 (.030) & .689 (.028) & .655 (.024)  & .683 (.022) & .690 (.026) & .672 (.025)\\
\texttt{[sc2-P-AT-M]}(MLPFL) & .664 (.028) & .671 (.025) & .652 (.019) & .701 (.026) & .693 (.028) & .674 (.023) & .712 (.031) & .686 (.027) & .690 (.029)\\
\texttt{[sc2-P-AV-B]} & .672 (.019) & .668 (.026) & .671 (.021) & \textbf{.734 (.034)} & .682 (.028) & .673 (.026) & .700 (.031) & .693 (.028) & .677 (.024)\\
\texttt{[sc2-P-AT-B]}(MLPFL) & \textbf{.693 (.029)} & \textbf{.682 (.016)} & \textbf{.684 (.017)} & .733 (.026) & \textbf{.700 (.028)} & \textbf{.685 (.014)} & \textbf{.723 (.020)} & \textbf{.710 (.026)} & \textbf{.691 (.014)}\\
\hline

\end{tabular}
}
\end{center}
% \vspace{-0.1in}
\label{tb:sc2_pp_g2}
\end{table*}

\begin{table*}[tp]
\caption{The knowledge tracing performance achieved by different approaches for each student subgroup categorized by age. The ``unspecified" subgroup encompasses students who haven't revealed their year of birth.}

\begin{center}

\scalebox{0.72}{

\begin{tabular}{ccccccccccccc}
\hline
\multicolumn{1}{c|}{Dataset}& \multicolumn{4}{c|}{FOC} & \multicolumn{4}{c|}{QDS}  & \multicolumn{4}{c}{MOSFETs}\\  \hline
\multicolumn{1}{c|}{Model} & $<80$   & $80-90$ & $>90$ & \multicolumn{1}{c|}{unspecified} &  $<80$   & $80-90$ & $>90$ & \multicolumn{1}{c|}{unspecified} & $<80$   & $80-90$ & $>90$ & unspecified\\ \hline\hline
\texttt{[sc2-L]} & .539 (.011) & .527 (.008) & .535 (.006) & .525 (.006) & .528 (.007) & .536 (.005) & .523 (.008) & .543 (.006) & .536 (.010) & .548 (.011) & .531 (.010) & .526 (.007) \\\hline
\texttt{[sc2-G]} & .551 (.015) & .569 (.015) & .557 (.011) & .561 (.009) & .563 (.013) & .561 (.008) & .541 (.010) & .556 (.009) & .547 (.014) & .558 (.011) & .562 (.016) & .544 (.007)
\\
\texttt{[sc2-G-AV-M]} & .548 (.011) & .541 (.013) & .537 (.015) & .526 (.005) & .531 (.011) & .528 (.010) & .530 (.011) & .526 (.013) & .523 (.007) & .562 (.011) & .528 (.016) & .557 (.011) \\
\texttt{[sc2-G-AT-M]}& .556 (.015) & .551 (.018) & .543 (.012) & .534 (.009) & .546 (.004) & .553 (.008) & .547 (.010) & .532 (.005) & .530 (.007) & .527 (.010) & .540 (.013) & .529 (.007) \\
\texttt{[sc2-G-AV-T]} & .546 (.010) & .534 (.008) & .581 (.010) & .537 (.013) & .549 (.008) & .530 (.005) & .541 (.013) & .534 (.004) & .538 (.009)  & .544 (.010) & .549 (.010) & .540 (.015)\\
\texttt{[sc2-G-AT-T]} & .537 (.009) & .549 (.012) & .562 (.015) & .546 (.007) & .521 (.006) & .538 (.010) & .551 (.013) & .556 (.011) & .554 (.014) & .549 (.011) & .533 (.017) & .537 (.009) \\
\hline
\texttt{FedIRT}& .603 (.012)& .592 (.017)& .622 (.008)& .629 (.011)& .631 (.013)& .602 (.007)& .599 (.011)& .607 (.009)& .629 (.015)& .611 (.019)& .628 (.009)& .618 (.010) \\
\texttt{[sc2-P-AV-M]} & .597 (.016) & .624 (.019) & .648 (.016) & .636 (.021) & .608 (.017) & .586 (.015) & .602 (.023) & .587 (.018) & .612 (.019) & .583 (.011) & .603 (.020) & .592 (.018)  \\
\texttt{[sc2-P-AT-M]}(MLPFL) & .619 (.025) & .657 (.025) & .662 (.022) & .648 (.027) & .613 (.019) & .607 (.015) & .618 (.015) & .600 (.009) & .628 (.016) & .617 (.011) & .632 (.017) & .629 (.019) \\
\texttt{[sc2-P-AV-B]} & .644 (.023) & .660 (.021) & \textbf{.680 (.025)} & .657 (.022) & .630 (.022) & .621 (.025) & .645 (.028) & .634 (.023) & .665 (.027) & .630 (.029) & .658 (.021) & .634 (.025) \\
\texttt{[sc2-P-AT-B]}(MLPFL) & \textbf{.654 (.022)} & \textbf{.672 (.031)} & .678 (.029) & \textbf{.662 (.027)} & \textbf{.648 (.025)} & \textbf{.632 (.021)} & \textbf{.658 (.028)} & \textbf{.643 (.019)} & \textbf{.672 (.027)} & \textbf{.648 (.025)} & \textbf{.669 (.028)} & \textbf{.651 (.027)} \\
\hline

\end{tabular}
}
\end{center}
% \vspace{-0.1in}
\label{tb:sc2_kt_a2}
\end{table*}

\begin{table*}[tp]
\caption{The outcome prediction performance achieved by various approaches for different student subgroups categorized by age. The ``unspecified" subgroup includes students who haven't provided their year of birth.}

\begin{center}

\scalebox{0.72}{

\begin{tabular}{ccccccccccccc}
\hline
\multicolumn{1}{c|}{Dataset}& \multicolumn{4}{c|}{FOC} & \multicolumn{4}{c|}{QDS}  & \multicolumn{4}{c}{MOSFETs}\\  \hline
\multicolumn{1}{c|}{Model} & $<80$   & $80-90$ & $>90$ & \multicolumn{1}{c|}{unspecified} & $<80$  & $80-90$ & $>90$ & \multicolumn{1}{c|}{unspecified} & $<80$   & $80-90$ & $>90$ & unspecified\\ \hline\hline
\texttt{[sc2-L]} & .582 (.013) & .600 (.019) & .619 (.017) & .553 (.008) & .617 (.020) & .588 (.018) & .601 (.023) & .554 (.008) & .559 (.021) & .593 (.015) & .561 (012.) & .592 (.018)\\\hline
\texttt{[sc2-G]} & .601 (.021) & .594 (.015) & .603 (.020) & .606 (.017) & .583 (.015) & .601 (.019) & .628 (.020) & .596 (.018) & .626 (.011) & .617 (.016) & .590 (.017) & .584 (.019)
\\
\texttt{[sc2-G-AV-M]} & .573 (.008) & .584 (.010) & .578 (.017) & .583 (.015) & .560 (.009) & .581 (.017) & .532 (.010) & .570 (.009) & .589 (.017) & .586 (.016) & .574 (.015) & .589 (.017) \\
\texttt{[sc2-G-AT-M]} & .600 (.020) & .592 (.015) & .597 (.011) & .602 (.018) & .581 (.015) & .596 (.013) & .553 (.009) & .582 (.011) & .603 (.018) & .597 (.017) & .581 (.011) & .613 (.015) \\
\texttt{[sc2-G-AV-T]} & .582 (.016) & .564 (.011) & .587 (.018) & .560 (.012) & .567 (.013) & .550 (.011) & .549 (.010) & .563 (.008) & .587 (.015)  & .560 (.017) & .573 (.011) & .568 (.015)\\
\texttt{[sc2-G-AT-T]} & .594 (.011) & .601 (.018) & .587 (.015) & .573 (.017) & .592 (.008) & .584 (.008) & .571 (.010) & .576 (.015) & .583 (.009) & .592 (.013) & .580 (.014) & .601 (.018) \\
\hline
\texttt{FedIRT} & .692 (.017) &	\textbf{.742 (.020)} &	.\textbf{760 (.027)} &		 .652 (.015) &	.709 (.016) &		\textbf{.728 (.018)} &	.701 (.009) &		.705 (.011) &		.693 (.013) &	.669 (.010) &		.624 (.012) &		.672 (.011)  \\
\texttt{[sc2-P-AV-M]} & .668 (.019) & .680 (.022) & .671 (.028) & .651 (.022) & .683 (.017) & .638 (.020) & .652 (.019) & .678 (.013) & .656 (.015) & .664 (.020) & .658 (.018) & .658 (.017)  \\
\texttt{[sc2-P-AT-M]}(MLPFL) & .673(.016) & .665 (.017) & .675 (.015) & .647 (.021) & .683 (.025) & .651 (.022) & .669 (.025) & .672 (.019) & .653 (.024) & .684 (.022) & .661 (.018) & .650 (.025) \\
\texttt{[sc2-P-AV-B]} & .690 (.022) & .704 (.029) & .706 (.027) & .683 (.025) & .692 (.028) & .700 (.019) & .719 (.021) & .708 (.019) & \textbf{.710 (.023)} & .714 (.020) & \textbf{.703 (.018)} & .661 (.021) \\
\texttt{[sc2-P-AT-B]}(MLPFL) & \textbf{.698 (.015)} & .713 (.019) & .706 (.023) & \textbf{.688 (.019)} & \textbf{.716 (.020)} & .694 (.018) & \textbf{.729 (.022)} & \textbf{.716 (.025)} & .708 (.027) & \textbf{.729 (.020)} & .698 (.020) & \textbf{.683 (.018)} \\
\hline

\end{tabular}
}
\end{center}
% \vspace{-0.1in}
\label{tb:sc2_pp_a2}
\end{table*}

\clearpage

% \begin{figure*}[!ht]
%     \centering
%     \setlength{\abovecaptionskip}{1mm}
%     \includegraphics[width=\linewidth]{figures/mean_var_info.pdf}
% 	 \caption{ 
%   Mean and standard deviation AUC scores of the personalized methods (\texttt{FedIRT}, \texttt{[sc2-P-AV-B]}, \texttt{[sc2-P-AT-B]}) across all subgroups, including the unspecified subgroup while modeling.
% } \label{fig:mean_var_noinfo}
% \end{figure*}

\end{document}